\documentclass[runningheads]{llncs}
\usepackage{eccv}

\usepackage{eccvabbrv}

\usepackage{graphicx}
\usepackage{xurl}
\usepackage{float}
\usepackage{booktabs}
\usepackage{longtable}
\usepackage{wrapfig}
\usepackage{caption}
\usepackage[T1]{fontenc}

\usepackage{hyperref}

\usepackage{orcidlink}

\usepackage{siunitx}
\newcommand{\nrLNFCodes}{140}          
\newcommand{\nrLabelsRaw}{78}          
\newcommand{\nrLabelsRawAg}{73}        
\newcommand{\nrLabelsExcluded}{8}      
\newcommand{\nrClasses}{70}            
\newcommand{\nrCropClasses}{65}        
\newcommand{\nrNonCropClasses}{5}      
\newcommand{\nrHierTop}{8}             
\newcommand{\nrHierGroups}{22}         
\newcommand{\nrHierInterm}{59}         

\newcommand{\topic}[1]{%
  \par\smallskip\noindent\textbf{\textit{#1}}\nobreak
}

\begin{document}

\title{SwissCrop25: A National Multi-Year Benchmark for Operational Crop Mapping}
\titlerunning{SwissCrop25}

\author{
Thomas Lauber\inst{1}\orcidlink{0000-0002-3118-432X} \and
Mehmet Ozgur Turkoglu\inst{1}\orcidlink{0000-0003-1446-2778} \and
Sélène Ledain\inst{1}\orcidlink{0009-0002-9321-1933} \and
Helge Aasen\inst{1}\orcidlink{0000-0003-4343-0476}}

\authorrunning{Lauber et al.}

\institute{Earth Observation of Agroecosystems Team, Agroscope, Switzerland \\
\email{thomas.lauber@agroscope.admin.ch, mehmet.tuerkoglu@agroscope.admin.ch, selene.ledain@agroscope.admin.ch, helge.aasen@agroscope.admin.ch}
}

\maketitle

\begin{abstract}
Operational crop mapping requires models that generalise across years, resolve fine-grained crop taxonomies, and distinguish cropland from surrounding landscapes. 
However, existing crop mapping data\-sets enable evaluation of these requirements only in isolation.
We therefore introduce SwissCrop25, a national-scale crop mapping benchmark dataset spanning seven growing seasons (2019--2025).
SwissCrop25 combines Sentinel-2 time series, daily temperature observations, a fine-grained \nrLabelsRawAg~crop taxonomy including grassland management types, and \nrNonCropClasses~explicit non-crop land cover classes.
To evaluate realistic deployment conditions, we define a leave-one-year-out protocol with joint cropland delineation and crop classification for benchmarking representative crop mapping architectures.
Evaluating U-TAE (convolutional temporal-attention model), TSViT (transformer-based spatio-temporal model), and Galileo (EO foundation model) reveals differences between architectures hidden by conventional benchmarks. 
In this setting, domain-specific models outperform Galileo, with TSViT achieving the best overall performance and a 12\,pp macro-mIoU advantage over U-TAE.
SwissCrop25 also exposes substantial interannual distribution shifts and shows that incorporating temperature-derived phenological information improves robustness.
Finally, in-season evaluation reveals a trade-off between models, with U-TAE performing better early in the season and TSViT gaining an advantage later through improved rare-class discrimination.
SwissCrop25 provides a challenging testbed for evaluating crop mapping systems under realistic operational conditions and is publicly released at 
\url{https://huggingface.co/datasets/EOA-team/SwissCrop25}.
\keywords{Crop type mapping \and Benchmark dataset \and Remote sensing \and Temporal generalisation \and Fine-grained classification}
\end{abstract}

\section{Introduction}
\label{sec:intro}

\begin{figure*}[t]
    \centering
    \includegraphics[width=\linewidth]{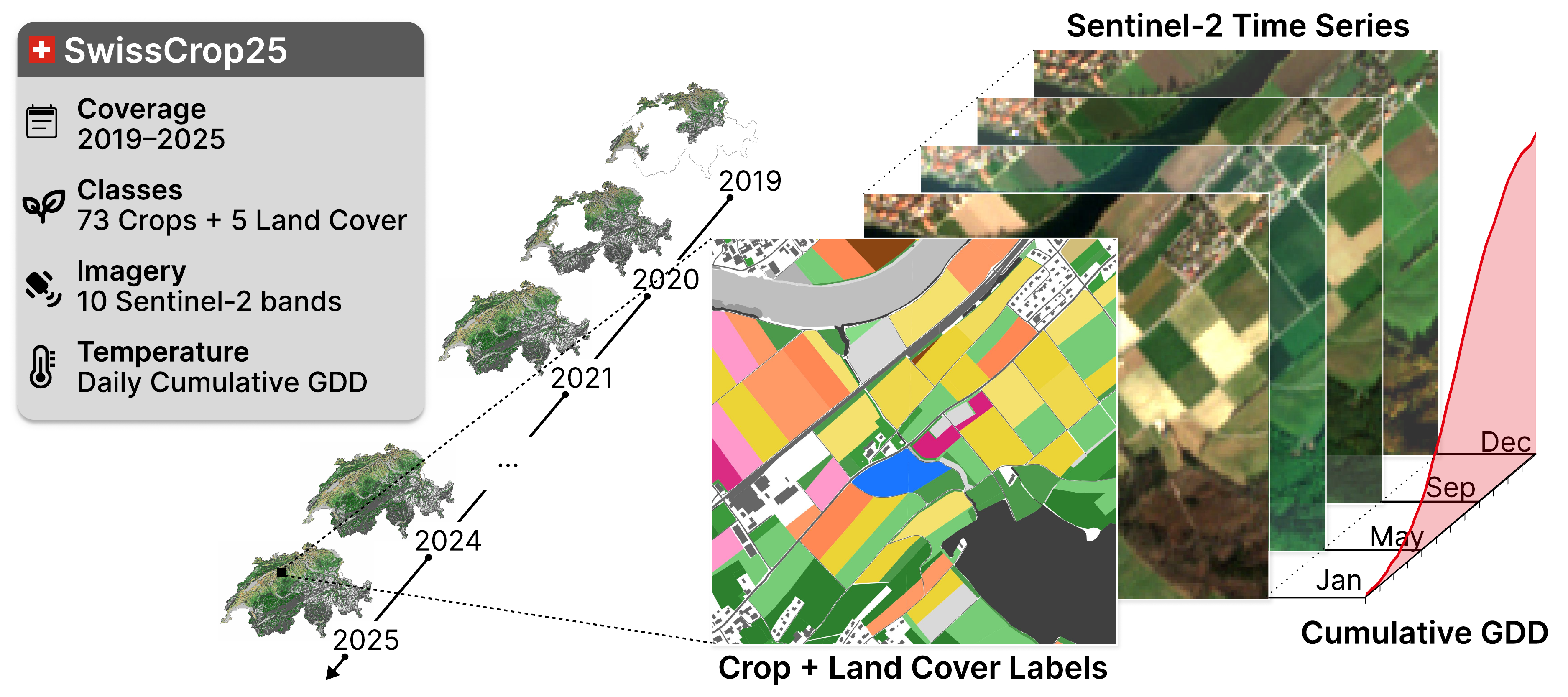}
    \caption{SwissCrop25 overview. Nationwide crop and land-cover labels paired with Sentinel-2 and cumulative growing degree day (GDD) time series, covering Switzerland across seven growing seasons (2019--2025).}
    \label{fig:dataset_overview}
\end{figure*}

Operational crop type mapping from satellite image time series has emerged as a foundational component of modern agricultural monitoring systems~\cite{becker-reshefCropTypeMaps2023}. 
It serves two distinct operational needs: (i)~annual crop inventories for national statistics, subsidy enforcement, carbon accounting, and environmental monitoring~\cite{atzbergerAdvancesRemoteSensing2013,sen4cap2021_validation}, and (ii)~in-season identification of emerging crops for operational decision-making, food-security systems and commodity markets~\cite{becker-reshefCropTypeMaps2023,wuChallengesOpportunitiesRemote2023}. 
The machine learning community has developed a wide range of methods for crop mapping, yet existing benchmarks do not fully reflect the requirements of operational crop mapping systems.
In practice, operational systems must generalise across years with different environmental conditions, distinguish long-tailed and fine-grained crop classes, operate without predefined cropland masks, and provide reliable predictions before the end of the growing season.
However, benchmark datasets that jointly support evaluation across these dimensions remain scarce. 
This paper addresses these requirements from the perspective of a national agricultural monitoring agency.

First, operational crop mapping models are typically trained on historical growing seasons and then deployed repeatedly in future years, where weather patterns, crop development, and management practices may differ from those observed during training~\cite{copernicuslandmonitoringserviceHighResolutionLayer2025,liAutomated10mResolution2026}. This makes operational deployment fundamentally a problem of temporal generalisation rather than interpolation within a single season. Robust evaluation should therefore test models under year-to-year distribution shifts, which is precisely what a strict Leave-One-Year-Out (LOYO) protocol provides. 
From an agronomic perspective, strong LOYO performance also suggests that a model captures crop phenology rather than merely recurring spectral-temporal patterns.
However, as we show in \Cref{sec:rw}, benchmark datasets that enable this form of evaluation remain scarce.

Second, crop class distributions in operational settings are highly imbalanced, with a few frequent crops dominating the data (e.g., wheat, maize, or grassland). Many benchmarks focus on a limited set of such broad and spectrally distinct classes. As a result, they primarily evaluate performance on relatively easy, common crops and provide limited insight into model behaviour under fine-grained classification and long-tailed class distributions. Robust performance is instead required on phenologically similar classes with different management implications (e.g., grain versus silage maize, or meadow classes of varying management intensities), as well as on rare but regionally or economically important crops (e.g., tobacco or hemp).

Third, most crop-type benchmarks assume a pre-defined agricultural mask, restricting evaluation to pixels already known to be cropland. However, crop mapping in operational settings is inherently a two-stage process: (i)~delineating cropland from non-cropland, and (ii)~assigning crop types within cropland. Ignoring the first task leads to inflated performance estimates, as errors on non-agricultural land are not penalised. In practice, cropland masks are often derived from external land-cover products~\cite{copernicuslandmonitoringserviceHighResolutionLayer2025}, introducing additional uncertainty that is typically not propagated into downstream classification. 

Finally, food security monitoring and commodity markets require crop type maps well before harvest, often when only early-season satellite observations are available. Yet, most benchmarks evaluate models only using complete growing-season time series, leaving open whether performance remains reliable when predictions must be made from truncated time series. End-of-season accuracy alone therefore cannot capture whether a model is suitable for applications where timely information is critical.

To address these four requirements, we introduce \textbf{SwissCrop25}, a country-wide crop mapping benchmark covering Switzerland over seven years (2019--2025; \cref{fig:dataset_overview}). 
While geographically limited to Switzerland, SwissCrop25 reflects the operational setting of national agricultural monitoring systems, where models are typically developed and deployed within a single country over multiple growing seasons. 
At the same time, Switzerland provides a diverse and challenging testbed, as its compact agro-climatic gradient and fragmented agricultural landscapes expose crop mapping models to substantial environmental and spatial variability over a relatively small geographic extent.
The dataset combines Sentinel-2 image time series with daily temperature observations, enabling ecophysiologically informed modelling and analysis of interannual variability. It covers \nrLabelsRawAg~crop classes, including fine-grained grassland management types absent from existing benchmarks, as well as \nrNonCropClasses~non-crop land cover classes for scene completeness. \Cref{tab:datasets} contrasts existing datasets along the dimensions above.

Our contributions are:
\begin{itemize}
\item \textbf{Dataset.} More than five years of national coverage enabling evaluation of temporal generalisation; \nrLabelsRawAg-class crop taxonomy including fine-grained grassland management types; \nrNonCropClasses~land cover classes for scene completeness; and daily temperature data enabling ecophysiologically informed modelling.
\item \textbf{Evaluation protocol.} A five-fold LOYO benchmark (2021--2025) with per-class reporting, explicit cropland mask evaluation, and in-season testing under truncated time series, each targeting a distinct operational gap.
\item \textbf{Baselines.} We benchmark three representative state-of-the-art model families: U-TAE (convolutional temporal-attention), TSViT (spatio-temporal transformer), and Galileo (Earth Observation (EO) foundation model), evaluated with and without ecophysiologically informed temporal modelling strategies.
\end{itemize}
\section{Related Work}
\label{sec:rw}

\topic{Operational crop mapping.} 
Satellite-based crop type mapping underpins continental and national monitoring systems for agricultural statistics, subsidy compliance, environmental reporting, food security, and forecasting~\cite{boryanMonitoringUSAgriculture2011,copernicuslandmonitoringserviceHighResolutionLayer2025,fisetteAAFCAnnualCrop2013,fisetteAnnualSpacebasedCrop2014,tettehAgriculturalLandUse2025,sen4cap2021_validation,hollandComplyingConservationCompliance2020,vantrichtWorldCerealDynamicOpensource2023}.
The diversity of these applications imposes requirements that go well beyond identifying dominant crops in a single year.
For instance, compliance monitoring under the European Union's Common Agricultural Policy and analogous programmes requires mapping agronomically relevant classes such as catch crops, nitrogen-fixing species, and fallow land~\cite{sen4cap2021_validation,hollandComplyingConservationCompliance2020}.
National greenhouse gas inventories require distinguishing management-intensity subclasses with substantially different emission factors, such as extensive versus intensive meadows~\cite{ipcc2006guidelines}.
Operational products covering full national territories must also distinguish cropland from non-agricultural land, since crop-type assignment first requires delineating cropland extent~\cite{boryanMonitoringUSAgriculture2011,copernicuslandmonitoringserviceHighResolutionLayer2025}.
Timeliness is a further requirement for public programmes~\cite{sen4cap2021_validation,liAutomated10mResolution2026} and commercial providers~\cite{senarasEarlySeasonCropClassification2024} alike, which increasingly provide rolling in-season products to support insurance applications, food-security alerts, and commodity markets that cannot wait for post-harvest estimates.
Across these applications, fine-grained taxonomies, full-scene coverage, and timely delivery emerge as recurring operational requirements.

\topic{Models for crop type mapping.}
Traditional operational crop mapping systems have relied on decision trees and random forests applied to engineered spectral-temporal features such as seasonal composites, vegetation indices, and phenological metrics~\cite{boryanMonitoringUSAgriculture2011,fisetteAAFCAnnualCrop2013,fisetteAnnualSpacebasedCrop2014,ghassemiEuropeanUnionCrop2024,deabelleyraArgentinaNationalMap2025}.
However, these engineered temporal summaries compress crop development trajectories into fixed descriptors, discarding much of their sequential structure.
Recurrent and convolutional models demonstrated the benefits of learning directly from these trajectories rather than relying on engineered temporal summaries~\cite{russwurmMultiTemporalLandCover2018,pelletierTemporalConvolutionalNeural2019,russwurmSelfattentionRawOptical2020}.
Attention-based approaches, notably L-TAE~\cite{garnotSatelliteImageTime2019} and U-TAE~\cite{faregarnotPanopticSegmentationSatellite2021}, further improved temporal modelling by learning to weight observations according to their phenological relevance while enabling parallel processing of the full time series.
TSViT~\cite{tarasiouViTsSITSVision2023} extended this idea with separate temporal and spatial attention stages and learnable crop-type class tokens, achieving state-of-the-art performance on existing crop mapping benchmarks.
More recently, Earth observation foundation models such as Presto~\cite{tsengLightweightPretrainedTransformers2024} and Galileo~\cite{pmlr-v267-tseng25a} are pre-trained on large multi-sensor archives to learn general-purpose representations, offering a potential route to reduced reliance on large labelled datasets.
These architectures are increasingly adopted in operational products, including the EU Copernicus crop type layer~\cite{copernicuslandmonitoringserviceHighResolutionLayer2025} and Germany's national mapping programmes~\cite{asamMappingCropTypes2022,phamTemporallyTransferableCrop2024}.
Despite these methodological advances, most models are evaluated on benchmarks that hold out data from the same year as training, leaving temporal generalisation largely untested.

\topic{Temporal generalisation in crop mapping.}
Growing seasons can vary substantially from year to year, with warm springs advancing green-up by weeks and droughts disrupting canopy development.
Consequently, models evaluated across years often show substantial accuracy drops relative to within-year evaluation~\cite{phamTemporallyTransferableCrop2024,wijesinghaEvaluatingSpatialTemporal2024}, highlighting the challenge of temporal generalisation.
Three broad strategies have been proposed to address this challenge.
First, data augmentation provides a model-agnostic approach, with Yuan~\etal~\cite{yuanEmpiricalStudyData2025} empirically evaluating eleven techniques including temporal shifting, dynamic time warping, and interpolation resampling under both same-year and cross-year settings.
Second, model-level adaptations aim to improve robustness to phenological misalignment. 
TimeMatch~\cite{nyborgTimeMatchUnsupervisedCrossregion2022} uses unsupervised domain adaptation to correct regional phenological shifts, Vincent~\etal~\cite{vincentPixelwiseAgriculturalImage2024} introduce temporal-shift invariance into prototype classifiers, and Nyborg~\etal~\cite{nyborgGeneralizedClassificationSatellite2022} replace calendar-date positional encodings with cumulative growing degree days to align equivalent growth stages.
Third, representation-level approaches address phenological misalignment before modelling. $T^{3}S$~\cite{turkogluT3SThinkThermal2026} re-indexes satellite observations according to cumulative growing degree days rather than calendar dates, aligning equivalent growth stages across years.
However, systematic evaluation of temporal generalisation under operational conditions requires multi-year national coverage, explicit LOYO splits, and environmental data capturing interannual phenological variability.

\topic{Crop type mapping datasets.}
Existing crop mapping benchmarks, spanning parcel-aggregated time series (TS) and satellite image time series (SITS), have driven substantial progress in model development but exhibit three structural limitations affecting their relevance for operational applications (\cref{tab:datasets}): insufficient temporal coverage, coarse class taxonomies, and incomplete scene coverage.

\begin{table}[t]
\centering
\caption{Comparison of crop type mapping benchmark datasets.
\emph{Type} distinguishes parcel-aggregated time series (TS) from satellite image time series (SITS).
\emph{Size} indicates the size of the uncompressed dataset.
\emph{LOYO} indicates whether leave-one-year-out evaluation is possible: \checkmark denotes repeated observations of the same spatial units across years, while (\checkmark) denotes multi-year coverage over different spatial units.
\emph{Temp} indicates availability of temperature data alongside imagery.
\emph{Mask} indicates explicit non-crop classes enabling evaluation beyond predefined cropland masks.}
\label{tab:datasets}
\setlength{\tabcolsep}{4pt}
\resizebox{1\linewidth}{!}{
\begin{tabular}{lllrllccc}
\toprule
\textbf{Dataset} & \textbf{Type} & \textbf{Scope} & \textbf{Size} & \textbf{Classes} & \textbf{\# Seasons} & \textbf{LOYO} & \textbf{Temp} & \textbf{Mask}\\
\midrule
BreizhCrops~\cite{russwurmBreizhCropsTimeSeries2019} &
TS &
Regional (FR) &
\SI{17}{\giga\byte} &
9 &
1\phantom{*} \hspace{0.3em}(2017) &
--- &
--- &
--- \\
TimeSen2Crop~\cite{weikmannTimeSen2CropMillionLabeled2021} &
TS   &
National (AT)  &
\SI{15}{\giga\byte} &
16     &
2\phantom{*} \hspace{0.3em}(2018--2019)     &
\checkmark          &
---          &
--- \\
EuroCropsML~\cite{reussEuroCropsMLTimeSeries2025}           &
TS   &
Transnational  &
\SI{4.5}{\giga\byte} &
176    &
1\phantom{*} \hspace{0.3em}(2021)           &
---          &
---          &
--- \\
TimeMatch~\cite{nyborgTimeMatchUnsupervisedCrossregion2022}         &
TS   &
Transnational  &
\SI{79}{\giga\byte} &
16     &
1\phantom{*} \hspace{0.3em}(2017)           &
---          &
\checkmark   &
--- \\
CropDeepTrans~\cite{barriereBoostingCropClassification2024}      &
TS   &
Transnational  &
\SI{102}{\giga\byte} &
151 &
5\phantom{*} \hspace{0.3em}(2016--2020)     &
\checkmark   &
---          &
--- \\
\midrule
MunichCrops~\cite{russwurmMultiTemporalLandCover2018}          &
SITS   &
Regional (DE)  &
\SI{39}{\giga\byte} &
17     &
2\phantom{*} \hspace{0.3em}(2016--2017)     &
(\checkmark) &
---          &
--- \\
PASTIS~\cite{faregarnotPanopticSegmentationSatellite2021}               &
SITS &
Regional (FR)  &
\SI{37}{\giga\byte} &
18     &
1\phantom{*} \hspace{0.3em}(2019)           &
---          &
---          &
\checkmark \\
ZueriCrop~\cite{turkogluCropMappingImage2021}          &
SITS & 
Regional (CH)  &
\SI{41}{\giga\byte} &
48     &
1\phantom{*} \hspace{0.3em}(2019)           &
---          &
---          &
--- \\
DENETHOR~\cite{kondmannDENETHORDynamicEarthNETDataset2021}         &
SITS &
Regional (DE)  &
\SI{255}{\giga\byte} &
9      &
2\phantom{*} \hspace{0.3em}(2018--2019)     &
\checkmark   &
---          &
--- \\
Sen4AgriNet~\cite{sykasSentinel2MultiyearMulticountry2022}      &
SITS &
Transnational  &
\SI{281}{\giga\byte}* &
158 &
2* \hspace{0.3em}(2016--2020)     &
(\checkmark) &
---          &
\checkmark \\
FLAIR-HUB~\cite{garioudFLAIRHUBLargescaleMultimodal2026}         &
SITS &
Regional (FR)  &
\SI{726}{\giga\byte} &
45     &
4\phantom{*} \hspace{0.3em}(2018--2021)     &
(\checkmark) &
---          &
\checkmark \\
\midrule
\textbf{SwissCrop25} &
\textbf{SITS} &
\textbf{National (CH)} &
\textbf{\SI{4.7}{\tera\byte}} &
\textbf{\nrLabelsRawAg} &
\textbf{7\phantom{*} (2019--2025)} &
\textbf{\checkmark} &
\textbf{\checkmark} &
\textbf{\checkmark} \\
\bottomrule
\multicolumn{8}{r}{\footnotesize $^*$only 2019--2020 are publicly available.} \\
\end{tabular}
}
\end{table}

\emph{Temporal coverage.}
Most SITS benchmarks cover a single growing season only~\cite{russwurmBreizhCropsTimeSeries2019,turkogluCropMappingImage2021,faregarnotPanopticSegmentationSatellite2021,reussEuroCropsMLTimeSeries2025}.
A few datasets span multiple years, but either provide limited replication over the same area~\cite{kondmannDENETHORDynamicEarthNETDataset2021,weikmannTimeSen2CropMillionLabeled2021} or confound temporal and spatial variation by assigning years to different regions~\cite{russwurmMultiTemporalLandCover2018,sykasSentinel2MultiyearMulticountry2022,nyborgTimeMatchUnsupervisedCrossregion2022,garioudFLAIRHUBLargescaleMultimodal2026}.
CropDeepTrans~\cite{barriereBoostingCropClassification2024} is the only benchmark enabling LOYO evaluation across multiple years, but its parcel-based design cannot be used for pixel-level evaluation. 
Moreover, no existing multi-year benchmark provides daily temperature observations needed to analyse phenological drivers of cross-year variation.

\emph{Class coverage.}
Many early benchmarks contain fewer than 20 classes, largely representing dominant crop types~\cite{russwurmBreizhCropsTimeSeries2019,russwurmMultiTemporalLandCover2018,faregarnotPanopticSegmentationSatellite2021,kondmannDENETHORDynamicEarthNETDataset2021}.
More recent datasets expand taxonomic detail, reaching up to 176 crop classes, with some capturing management distinctions such as silage versus grain maize~\cite{turkogluCropMappingImage2021,reussEuroCropsMLTimeSeries2025,barriereBoostingCropClassification2024,sykasSentinel2MultiyearMulticountry2022}.
However, grasslands remain poorly resolved, with ZueriCrop~\cite{turkogluCropMappingImage2021} providing the most detailed available classification by separating meadow, pasture, and biodiversity areas. 
Yet, no existing dataset captures operationally relevant intensity variants within these grassland categories. 

\emph{Scene completeness.}
Most crop-type benchmarks restrict evaluation to a predefined cropland mask.
Consequently, models are not penalised for confidently labelling forests or urban pixels as crops, producing overly optimistic accuracy estimates.
Only a small number of datasets include non-crop classes alongside crops~\cite{faregarnotPanopticSegmentationSatellite2021,turkogluCropMappingImage2021,sykasSentinel2MultiyearMulticountry2022,garioudFLAIRHUBLargescaleMultimodal2026}, but these are typically based on Land Parcel Identification System (LPIS) information rather than independent land-cover mapping.
However, LPIS-derived non-crop labels originate from declared parcels and may therefore absorb agricultural areas outside declared parcels into the background class.
No existing benchmark provides explicitly defined non-crop classes based on independent land-cover information.
\section{The SwissCrop25 Dataset}
\label{sec:dataset}

SwissCrop25 is a national crop inventory benchmark derived from Switzerland's operational agricultural reporting system, covering the full territory (\SI{41285}{\kilo\meter\squared}) across seven growing seasons (2019--2025). 
Crop labels consist of \nrLabelsRawAg{} agricultural classes structured in an agronomic hierarchy, together with \nrNonCropClasses{} non-crop land cover classes (forest, water, built-up, unproductive land, and wetland). The dataset pairs these labels with Sentinel-2 image time series and daily temperature observations.

\subsection{Crop Type Labels}
\label{sec:dataset:labels}

\topic{Parcel-level annotations.}
Crop type labels are derived from the Agricultural Cultivated Areas dataset (Landwirtschaftliche Nutzfl\"{a}che, LNF), compiled from cantonal (Swiss administrative region) land-use declarations and published annually by the Swiss Federal Office for Agriculture (FOAG)~\cite{foag2024_kulturflaechen,geodienste_nutzungsflaechen}. The LNF is the Swiss equivalent of the Land Parcel Identification System (LPIS) used across the European Union, assigning a single crop or land-use code to each parcel-year record. Each parcel is declared according to its intended primary crop, defined as the crop occupying the land for the longest period during the growing season and established by 1~June. Because declarations reflect planting intent rather than confirmed end-of-season outcomes, the recorded code may differ from realised agricultural use. This introduces label noise that is inherent to administrative LPIS-based datasets, most notably for crops whose final use is determined at harvest, such as silage versus grain maize. Quantifying this uncertainty would require independent harvest observations or field inspections, which are not available at national scale.

To ensure spatial consistency despite incomplete national coverage in 2019 and 2020, we exclude canton-year combinations whose mapped LNF extent falls below 90\% of the canton's maximum extent observed across the full time series. This affects nine cantons in 2019 and three in 2020. We additionally remove duplicate parcel entries from the canton of Ticino in 2021. Coverage is complete across all cantons from 2021 onward (Supp.~\cref{supp:tab:dataset_stats}).

\topic{Land cover supplement.}
We supplement the LNF with \nrNonCropClasses{} land cover classes from the national topographic landscape model (swissTLM3D)~\cite{swisstopo_swisstlm3d}: forest, water, built-up, unproductive land, and wetland. These classes provide explicit non-crop labels, enabling complete land-cover representation rather than treating non-agricultural areas as unlabelled background. They are also used to refine alpine pasture polygons, as they are not delineated against adjacent land-cover types, such as forest and unproductive land, in the LNF. 
Outside alpine pasture areas, the LNF remains the primary source of agricultural labels where LNF and swissTLM3D overlap (details in Supp.~\cref{supp:preprocessing}).
The \nrNonCropClasses{} non-crop classes are part of the full \nrLabelsRaw{}-class label space. The combined polygon data are rasterised at \SI{10}{\meter} using a fractional coverage approach~\cite{bastonExactextractrFastExtraction2024} rather than the centre-of-pixel rule, which better preserves narrow field strips and small parcels (details in Supp.~\cref{supp:preprocessing}).

\topic{Class hierarchy.}
We aggregated the \nrLNFCodes{} agricultural LNF codes into \nrLabelsRaw{} labels (\nrLabelsRawAg{} agricultural and \nrNonCropClasses{} land cover). Agricultural labels preserve distinctions in management intensity (e.g., meadow intensity classes), declared end use (e.g., Grain Maize and Silage Maize), and minority crops (e.g., Tobacco). The classes are organised into a four-level hierarchy comprising \nrLabelsRaw{} output labels, \nrHierInterm{} intermediate classes, \nrHierGroups{} crop-type groups, and \nrHierTop{} top-level categories. All output classes are additionally mapped to HCAT4~\cite{claverieEuroCropsV20Multiannual2026}, which enables compatibility with EuroCrops~\cite{schneiderEuroCropsLargestHarmonized2023} and the European Union's LPIS taxonomy.

\topic{Class distribution.}
The resulting dataset exhibits a pronounced long-tail distribution spanning nearly five orders of magnitude, from Intensive Meadow, Ley and Winter Wheat to minority crops such as Safflower and Special Crops (Supp.~\cref{fig:label_dist}). Class imbalance ratios exceed 180{,}000:1, which reflects the actual structure of Swiss agriculture and is intentionally preserved, maintaining the challenge of recognising rare crop types rather than restricting evaluation to dominant classes.

\subsection{Sentinel-2 Image Time Series}
\label{sec:dataset:s2}

Sentinel-2A and Sentinel-2B provide multispectral observations with revisit intervals of approximately 2--3 days over Switzerland~\cite{druschSentinel2ESAsOptical2012}. We use the 10 Sentinel-2 spectral bands covering visible, red-edge, near-infrared, and shortwave infrared regions across the 2019--2025 period, resampling the native \SI{20}{\meter} bands to a common \SI{10}{\meter} grid. Imagery is downloaded from Microsoft Planetary Computer~\cite{microsoftopensourceMicrosoftPlanetaryComputerOctober2022}, BRDF-corrected~\cite{monteroFacilitatingAdvancedSentinel22024}, and organised into spatio-temporal data cubes. Each cube represents one year and consists of $128 \times 128$ pixels (\SI{1280}{\meter}$\times$\SI{1280}{\meter}) aligned to UTM zone 32N on a common \SI{10}{\meter} grid. Each observation is accompanied by two cloud masks: the native Sentinel-2 cloud flag and the multi-class CloudSEN12+ score~\cite{aybarCloudSEN12LargestDataset2024}, which support flexible cloud filtering strategies.

\subsection{Temperature Data}
\label{sec:dataset:temp}

\begin{figure}[!b]
  \centering
  \includegraphics[width=\linewidth]{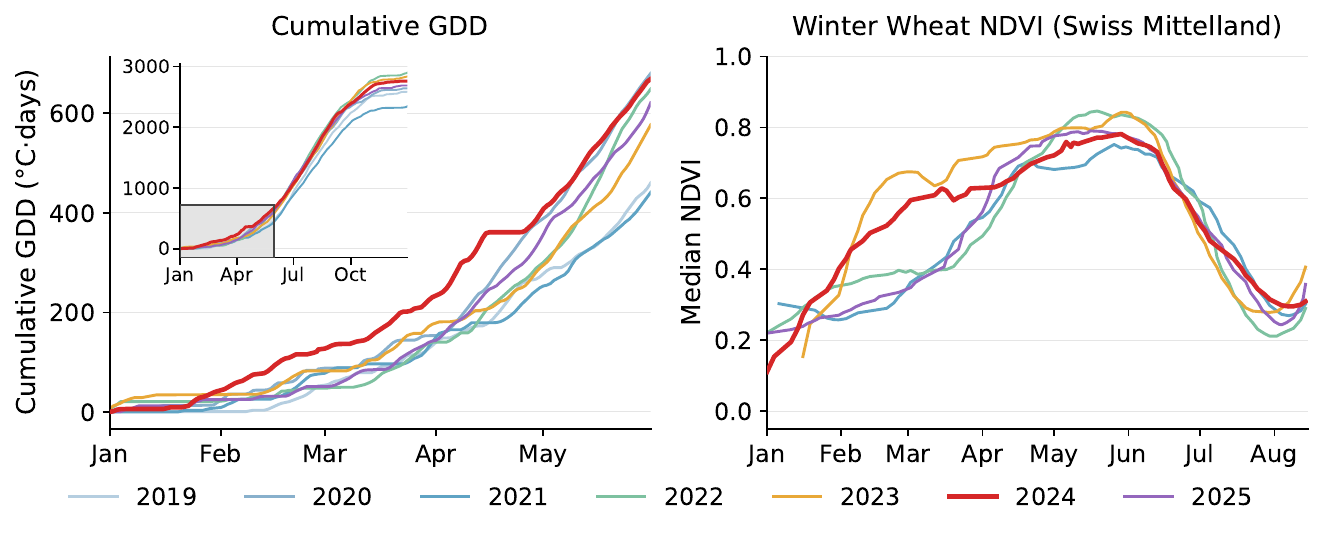}
  \caption{Interannual weather and growth variability.
    \textbf{Left:} Cumulative GDD averaged across Switzerland from 1~January
    through 31~May; inset shows the full year.
    \textbf{Right:} Median NDVI of winter wheat pixels in the Swiss Mittelland region around Bern.
    2019 and 2020 are absent from the NDVI panel due to incomplete parcel data.}
  \label{fig:gdd_ndvi}
\end{figure}

SwissCrop25 includes a per-cube time series of cumulative growing degree days (GDD) derived from MeteoSwiss gridded daily mean temperature~\cite{meteoswiss2021_tabsd}. 
Since crop identity is unknown at inference time, we use a crop-agnostic base temperature of $T_{\text{base}} = 0\,^{\circ}\mathrm{C}$ rather than crop-specific base temperatures, which would require prior knowledge of the target crop. For each cube and day, the \SI{1}{\kilo\meter} resolution daily mean temperature $T_{\text{mean}}$ is averaged over the cube extent and accumulated as $\max\!\bigl(T_{\text{mean}} - T_{\text{base}},\,0\bigr)$ from 1 January, yielding a cumulative GDD time series for each cube.

The seven-year span captures meaningful interannual weather variability. For example, anomalously warm winter temperatures in 2024 led to pronounced early spring development signals in winter cereals (+53\% cumulative GDD by 1 April, accompanied by earlier winter wheat NDVI development; \cref{fig:gdd_ndvi}). SwissCrop25 therefore provides a challenging test case for temporal generalisation across years with distinct phenological trajectories.

\subsection{Data Availability}
\label{sec:dataset:availability}

SwissCrop25 is publicly available on Hugging Face: \url{https://huggingface.co/datasets/EOA-team/SwissCrop25}.
The dataset is released under the Creative Commons Attribution 4.0 International licence (CC BY 4.0).
Code for dataset generation, preprocessing, and reproduction of all benchmark results is available at \url{https://github.com/thomaslauber/SwissCrop25}.

\section{Experiments}
\label{sec:experiments}

SwissCrop25 is designed to evaluate crop mapping systems under operational deployment conditions.
We structure the experiments around four key challenges that existing datasets typically address only in isolation: (i)~scene completeness, combining cropland delineation and crop type mapping; (ii)~temporal generalisation, using a five-year leave-one-year-out (LOYO) protocol spanning genuine interannual weather variability; (iii)~taxonomic depth, with \nrCropClasses-class fine-grained crop classification; and (iv) in-season usability through early-season evaluation.
We further assess low-resource performance and computational efficiency.
Together, these experiments test whether models can accurately delineate cropland, classify fine-grained crop classes, generalise across years, and deliver reliable in-season predictions.

\topic{Setup.}
We evaluate U-TAE~\cite{faregarnotPanopticSegmentationSatellite2021}, TSViT~\cite{tarasiouViTsSITSVision2023}, and Galileo~\cite{pmlr-v267-tseng25a}, representing established crop-mapping architectures and a recent EO foundation model.
Galileo is evaluated in both fine-tuned and frozen-encoder configurations.
Each LOYO split holds out one calendar year as the test set (2021--2025 in turn), uses the immediately preceding year for validation to avoid temporal leakage, and trains on all remaining years. 
The two partial years (2019--2020) are included as training data only due to their incomplete national coverage.
Additional naive temporal baselines exploiting the LOYO structure are provided in Supp.~\cref{supp:extended,supp:temporal}.
While the dataset includes parcel-level location metadata enabling geographic cross-validation, spatial locations are intentionally reused across years, mirroring operational deployment where models are repeatedly applied to the same agricultural landscape.
Reported performance therefore reflects year-to-year generalisation under realistic deployment conditions rather than transfer to unseen geography.
We additionally evaluate two temporal representation strategies designed to improve interannual generalisation: phenology-aware Growing Degree Day (GDD)-bin sampling ($T^{3}S$)~\cite{turkogluT3SThinkThermal2026} and sinusoidal thermal positional encoding (TPE)~\cite{nyborgGeneralizedClassificationSatellite2022} (\cref{sec:exp:temporal}).
Unless otherwise stated, all reported results use the best-performing configuration per architecture: $T^{3}S$\,+\,TPE for U-TAE and TSViT, and $T^{3}S$ alone for Galileo-nano (whose pretrained positional encoding is fixed).

Of the \nrLabelsRaw{} labels in the class hierarchy, \nrClasses{} are retained for training and evaluation.
The remaining \nrLabelsExcluded{} are excluded due to insufficient temporal coverage, low sample counts, ambiguous mixed-culture definitions, or the absence of a meaningful land-cover class.
All models are trained on all \nrClasses\ classes and evaluated in two stages: first as a binary cropland mask (agricultural vs.\ non-agricultural), then on the \nrCropClasses\ crop classes only.
All classification metrics---OA, GIoU (both micro), mIoU, and mF1 (both macro)---are computed over crop classes only, while calibration metrics (Expected Calibration Error (ECE) and NLL) are computed across all classes. 
For definitions of the metrics, see Supp.~\cref{supp:metrics}.
All models are trained with class-balanced cross-entropy loss~\cite{cuiClassBalancedLossBased2019a}; full training details and hyperparameters are reported in Supp.~\cref{supp:training}.

\begin{wraptable}{r}{0.38\linewidth}
\vspace{-30pt}
\centering
\caption{Agricultural mask evaluation (IoU$_\mathrm{ag}$), averaged over five LOYO splits. Full precision, recall, and F1 in Supp.~\cref{supp:tab:mask_full}. Best is \textbf{bold}.}
\label{tab:mask}
\setlength{\tabcolsep}{6pt}
\begin{tabular}{lr}
\toprule
\textbf{Model} & \textbf{IoU$_\mathrm{ag}$ (\%) $\uparrow$} \\
\midrule
U-TAE & \textbf{89.2} \\
TSViT & 88.9 \\
Galileo-nano & 86.1\\
\bottomrule
\end{tabular}
\vspace{-15pt}
\end{wraptable}

\subsection{Scene Completeness}
\label{sec:exp:mask}

Models trained jointly on crop and land cover classes achieve only 86--89\% of agricultural area (IoU$_\text{ag}$, \cref{tab:mask}), indicating a systematic shortfall overlooked in benchmarks that assume a perfect cropland mask.
U-TAE and TSViT achieve comparable IoU$_\text{ag}$ (89.2\% vs.\ 88.9\%), with Galileo-nano lower at 86.1\%.
The errors are predominantly false negatives, meaning all fine-tuned models achieve high precision (>96\%) but only 89--92\% recall, missing 8--11\% of cropland area. These omissions are concentrated in forest-adjacent classes (e.g.\ Forest Pasture, Chestnut Orchards) and semi-natural grasslands (e.g.\ Extensive Meadow, Alpine Pasture), which are frequently confused with unproductive land due to their spectral similarity.
Frozen encoder variants show slightly lower recall (85--87\%), suggesting that task-specific adaptation provides a modest benefit for recovering agricultural areas. Full precision, recall, and F1 statistics and per-model confusion matrices are reported in Supp.~\cref{supp:tab:mask_full} and \cref{supp:classification}, respectively.

\subsection{Temporal Generalisation}
\label{sec:exp:temporal}

By evaluating models across multiple growing seasons, the LOYO protocol exposes temporal failure modes that remain invisible in conventional single-year benchmarks.
\Cref{tab:results} shows that U-TAE and TSViT achieve near-identical overall accuracy (77.7 vs.\ 77.1\%) and GIoU (63.5 vs.\ 62.7\%), yet mIoU and mF1 reveal a 12\,pp gap (35.8 vs.\ 48.1\%) and 15\,pp gap (45.7 vs.\ 60.7\%), respectively, differences substantially larger than the observed variations across LOYO splits.
Fine-tuned Galileo-nano (30.4\% mIoU) falls behind the other models, while frozen variants (Galileo-nano: 14.1\%; Galileo-base: 20.6\%; full results in Supp.~\cref{supp:tab:results_full}) perform substantially worse, indicating that the evaluated pretrained representations alone are insufficient for competitive crop mapping in this setting, even with a larger backbone (Galileo-base).
U-TAE achieves better calibration than TSViT (ECE 0.77\% vs.\ 1.86\%), highlighting that accuracy and confidence reliability represent distinct operational objectives.

\begin{table*}[t]
\centering
\caption{LOYO benchmark results. U-TAE and TSViT use $T^{3}S$\,+\,TPE; Galileo is the full fine-tuned nano model and uses $T^{3}S$ only (fixed pretrained PE). Frozen encoder variants are reported in Supp.~\cref{supp:tab:results_full}.
Classification metrics (OA, GIoU, mIoU, mF1) are computed over crop classes only, while calibration metrics (ECE, NLL) use the full label space; best value per split and metric is \textbf{bold}.}
\label{tab:results}
\setlength{\tabcolsep}{5pt}
\resizebox{\linewidth}{!}{
\begin{tabular}{llrrrrrr}
\toprule
\textbf{Test year} & \textbf{Model} & \textbf{OA (\%) $\uparrow$} & \textbf{GIoU (\%) $\uparrow$} & \textbf{mIoU (\%) $\uparrow$} & \textbf{mF1 (\%) $\uparrow$} & \textbf{ECE (\%) $\downarrow$} & \textbf{NLL\,(-) $\downarrow$} \\
\midrule
2021 & U-TAE & \textbf{77.4} & \textbf{63.1} & 36.5 & 46.5 & \textbf{0.99} & \textbf{0.39} \\
 & TSViT & 76.9 & 62.5 & \textbf{46.6} & \textbf{58.9} & 2.36 & 0.41 \\
 & Galileo-nano (FT) & 73.0 & 57.5 & 30.5 & 41.3 & 1.06 & 0.50 \\
\midrule
2022 & U-TAE & \textbf{79.2} & \textbf{65.5} & 37.6 & 47.4 & \textbf{0.12} & \textbf{0.36} \\
 & TSViT & 78.7 & 64.9 & \textbf{50.0} & \textbf{62.8} & 0.13 & \textbf{0.36} \\
 & Galileo-nano (FT) & 73.8 & 58.5 & 31.3 & 42.2 & 0.55 & 0.48 \\
\midrule
2023 & U-TAE & \textbf{78.2} & \textbf{64.2} & 33.7 & 43.1 & \textbf{0.35} & \textbf{0.38} \\
 & TSViT & 77.0 & 62.6 & \textbf{47.9} & \textbf{60.6} & 1.71 & 0.40 \\
 & Galileo-nano (FT) & 72.2 & 56.5 & 28.3 & 38.6 & 0.65 & 0.50 \\
\midrule
2024 & U-TAE & \textbf{75.3} & \textbf{60.4} & 33.2 & 43.2 & 1.09 & \textbf{0.44} \\
 & TSViT & \textbf{75.3} & \textbf{60.4} & \textbf{44.6} & \textbf{57.7} & 2.48 & 0.45 \\
 & Galileo-nano (FT) & 71.0 & 55.1 & 28.3 & 38.9 & \textbf{1.00} & 0.54 \\
\midrule
2025 & U-TAE & \textbf{78.4} & \textbf{64.4} & 38.3 & 48.3 & 1.29 & \textbf{0.40} \\
 & TSViT & 77.6 & 63.4 & \textbf{51.3} & \textbf{63.6} & 2.60 & 0.42 \\
 & Galileo-nano (FT) & 74.4 & 59.2 & 33.6 & 44.6 & \textbf{0.34} & 0.49 \\
\midrule
\midrule
Mean $\pm$ Std & U-TAE & \textbf{77.7} $\pm$ 1.5 & \textbf{63.5} $\pm$ 2.0 & 35.8 $\pm$ 2.3 & 45.7 $\pm$ 2.4 & 0.77 $\pm$ 0.50 & \textbf{0.40} $\pm$ 0.03 \\
 & TSViT & 77.1 $\pm$ 1.2 & 62.7 $\pm$ 1.6 & \textbf{48.1} $\pm$ 2.7 & \textbf{60.7} $\pm$ 2.5 & 1.86 $\pm$ 1.03 & 0.41 $\pm$ 0.03 \\
 & Galileo-nano (FT) & 72.9 $\pm$ 1.3 & 57.4 $\pm$ 1.6 & 30.4 $\pm$ 2.2 & 41.1 $\pm$ 2.5 & \textbf{0.72} $\pm$ 0.31 & 0.50 $\pm$ 0.02\\
\bottomrule
\end{tabular}
}
\end{table*}

Interannual variability is substantial, with TSViT mIoU ranging from 44.6\% (2024) to 51.3\% (2025).
The 2024 split also shows the highest NLL across all models (U-TAE 0.44, TSViT 0.45, Galileo-nano 0.54), with ECE also elevated (U-TAE 1.09\%, TSViT 2.48\%, Galileo-nano 1.00\%).
The 2024 split represents the most challenging year, with anomalously warm winter temperatures associated with shifts in phenological signals (\cref{fig:gdd_ndvi}). 

\topic{Phenological alignment.}
To assess whether temporal misalignment contributes to these year-specific degradations, we next evaluate phenological alignment strategies.
\Cref{tab:temporal_encoding} evaluates two interventions against a day-of-year (DOY) temporal alignment baseline with cloud filtering: model-agnostic thermal time-based temporal subsampling via $T^{3}S$~\cite{turkogluT3SThinkThermal2026} and sinusoidal thermal positional encoding (TPE)~\cite{nyborgGeneralizedClassificationSatellite2022}.
In terms of mIoU, $T^{3}S$ yields modest, architecture-dependent gains: $+$1.4\,pp for TSViT and $+$1.6\,pp for Galileo-nano. U-TAE shows no consistent benefit.
TPE provides the largest improvement, increasing TSViT by $+$10.7\,pp over $T^{3}S$ alone and U-TAE by $+$1.2\,pp.
TSViT benefits from both interventions, whereas U-TAE shows only a modest response to TPE and no consistent gain from $T^{3}S$.
The asymmetric response is consistent with differences in temporal representation: TSViT's class-token attention benefits from both improved phenological sampling and explicit temporal indexing, whereas U-TAE mainly benefits from positional encoding, suggesting that its temporal attention already compensates for irregular observation timing.
Full per-split results are in Supp.~\cref{supp:tab:temporal_encoding}.
For TSViT in 2024, TPE substantially reduces confusion among minority winter cereals, recovering Triticale and Rye to above 70\% recall (Supp.~\cref{supp:fig:confmat_winter_cereals_2024}).
Despite these improvements, 2024 remains the most challenging LOYO split across all encoding strategies, indicating that while temperature-driven phenological shifts explain part of the observed errors, they do not fully account for the overall degradation in performance.

\begin{table}[t]
\centering
\caption{Effect of thermal time sampling and positional encoding on crop mIoU (\%).
Baseline uses 24 least-cloudy observations sampled uniformly in calendar time. 
$T^{3}S$ samples uniformly in thermal time (cumulative GDD)~\cite{turkogluT3SThinkThermal2026}.
Thermal PE adds sinusoidal thermal positional encoding~\cite{nyborgGeneralizedClassificationSatellite2022}. 
Galileo-nano cannot use Thermal PE because its pretrained positional encoding is month-based. Values are means over five LOYO splits; full results in Supp.~\cref{supp:tab:temporal_encoding}.}
\label{tab:temporal_encoding}
\setlength{\tabcolsep}{6pt}
\begin{tabular}{lccc}
\toprule
\textbf{Model} & \textbf{Baseline} & \textbf{$\boldsymbol{T^{3}S}$} & \textbf{$\boldsymbol{T^{3}S}$\,+\,Thermal PE} \\
\midrule
U-TAE & 34.7 & 34.6 & \textbf{35.8} \\
TSViT & 36.0 & 37.4 & \textbf{48.1} \\
Galileo-nano (FT) & 28.8 & \textbf{30.4} & ---\\
\bottomrule
\end{tabular}
\end{table}

\subsection{Fine-grained Classification}
\label{sec:exp:classification}

\topic{Taxonomic granularity.}
The hierarchical label structure of SwissCrop25 enables evaluation at multiple semantic resolutions.
Supp.~\cref{supp:fig:taxonomy} shows crop mIoU and mF1 as a function of taxonomy level, from 3 coarse land-use categories (lv3: arable, grassland, and permanent) down to the full \nrCropClasses-class leaf taxonomy.
At lv3, all three models score within 5\,pp of each other in mIoU, suggesting that broad land-use categories require less specialised representations.
As taxonomic specificity increases, the curves diverge sharply and the mIoU gap between TSViT and U-TAE grows from under 1\,pp at lv3 to 12\,pp at leaf level, demonstrating that model rankings depend strongly on the semantic resolution of evaluation; full per-class and hierarchical IoU values are reported in Supp.~\cref{supp:tab:perclass}.

\topic{Long-tail difficulty.}
Within the leaf taxonomy, performance differences are concentrated among rare classes. 
Supp.~\cref{supp:fig:longtail} shows crop mIoU restricted to increasingly rare classes (by frequency percentile), confirming that TSViT increasingly outperforms U-TAE as evaluation is restricted to rarer classes.
TSViT's per-class tokens, which enable class-specific global aggregation over image patches, may contribute to this advantage.
This is consistent with the stronger response of TSViT to temporal representation choices (\cref{sec:exp:temporal}), where class-specific tokenisation may help capture subtle phenological signatures of rare crops that are otherwise difficult to learn under strong class imbalance.

\topic{Grassland classes.}
Among the grassland subclasses newly introduced in SwissCrop25, intensive and extensive management types are reliably distinguished with high recall (Intensive Meadow: $\sim$68\%; Extensive Meadow: $\sim$52\%), while Less Intensive Meadow is difficult to classify for all models ($\sim$12\%), consistent with its intermediate position between the two management categories (Supp.~\cref{supp:fig:confmat_utae}).
Forest Pasture is a notable exception to the general TSViT advantage on minority classes. U-TAE substantially outperforms TSViT (42\% vs.\ 24\%), highlighting the importance of spatial context for certain crop types. Forest Pasture likely benefits from neighbourhood context, as forest proximity provides a strong spatial cue that may be better captured by U-TAE's multi-scale convolutions than TSViT's attention-based spatial aggregation.

\subsection{In-season Usability}
\label{sec:exp:inseason}

End-of-season accuracy does not fully capture the operational value of a crop mapping system, as many applications require predictions before the end of the growing season. 
We therefore evaluate full-season-trained models using progressively longer portions of the annual time series, truncating observations after each calendar month from January to December.
This mimics operational deployment, where models trained on historical full-season data are applied mid-season without retraining; see Supp.~\cref{supp:inseason_protocol} for implementation details.

\begin{figure}[t]
  \centering
  \includegraphics[width=\linewidth]{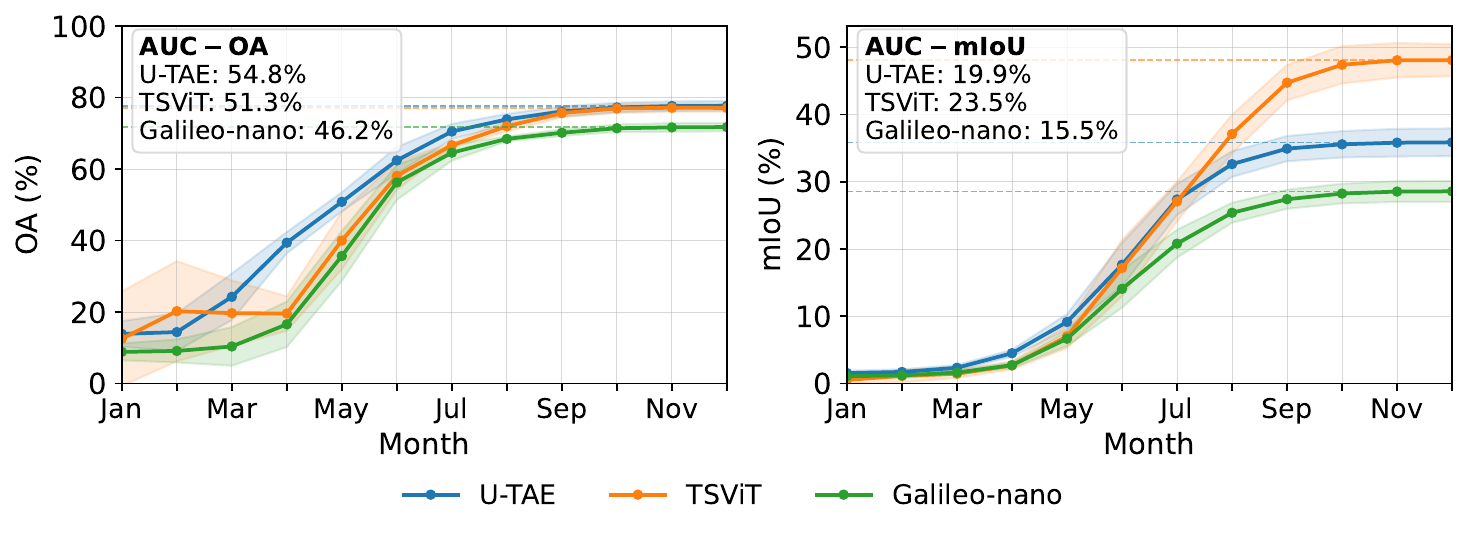}
  \caption{In-season OA (left) and mIoU (right) as a function of month cutoff (mean ±1 std across five LOYO splits). 
  U-TAE leads early OA, whereas TSViT gains a late-season advantage in mIoU.
  }
  \label{fig:inseason}
\end{figure}

\Cref{fig:inseason} shows that model ranking depends on both evaluation metric and prediction timing.
For OA, U-TAE leads early in the season, but TSViT progressively closes the gap and reaches comparable performance by the end of the season.
For mIoU, the trajectories diverge more strongly, with U-TAE initially leading, but TSViT overtaking in August and gaining its largest advantage among rare classes.
We further summarise in-season performance using the area under the in-season performance curve (AUC), integrating each metric across the twelve monthly cutoff points.
Although U-TAE and TSViT reach similar end-of-season OA, U-TAE achieves higher AUC-OA (54.8\% vs.\ 51.3\%), reflecting its stronger performance earlier in the season.
In contrast, TSViT achieves higher AUC-mIoU (23.5\% vs.\ 19.9\%), driven by its stronger late-season improvements in fine-grained crop classification.

The divergence between OA and mIoU reflects differences in class frequency.
U-TAE maintains an advantage on common crops early in the season, whereas TSViT progressively improves rare-class discrimination as additional observations become available, leading to its larger end-of-season mIoU advantage across the crop taxonomy (Supp.~\cref{supp:fig:inseason_classfreq}). These results show that the preferred architecture depends on the deployment objective: U-TAE may be advantageous for earlier predictions, whereas TSViT provides greater value when later-season fine-grained classification is required.

\subsection{Efficiency and Scalability}
\label{sec:exp:efficiency}

We evaluate data efficiency by training models on a randomly sampled 10\% subset of training cubes (Supp.~\cref{supp:training}); results are reported in Supp.~\cref{supp:tab:lowresource}.
U-TAE shows the largest degradation, retaining 49\% of full-data mIoU, compared with 64\% for TSViT and 61\% for Galileo-nano.
U-TAE's calibration deteriorates most strongly, with ECE increasing from 0.77\% to 18.9\%, whereas TSViT (1.5\%) and Galileo-nano (0.7\%) remain stable.

Computational costs are reported in Supp.~\cref{supp:tab:compute}. TSViT achieves the highest accuracy at 35\% higher training cost than U-TAE, while being slightly faster at inference.
Fine-tuned Galileo-nano incurs $3\times$ higher training cost than TSViT with substantially lower performance, while Galileo-base is impractical to fine-tune even on 4 GH200 GPUs and requires $5\times$ longer inference than TSViT.
Overall, dedicated crop mapping architectures provide a stronger accuracy--efficiency trade-off than the evaluated pretrained models, supporting their use in operational-scale crop mapping.
\section{Conclusion}
\label{sec:conclusion}

We introduced SwissCrop25, a national-scale benchmark for evaluating crop mapping systems under realistic operational conditions. 
The dataset combines seven years of Sentinel-2 observations, daily temperature data, a fine-grained crop taxonomy including grassland management types, and explicit non-crop land cover classes. 
Together with leave-one-year-out and in-season evaluation frameworks, it enables systematic assessment of scene completeness, temporal generalisation, semantic granularity, and prediction timing.
Our results show that model rankings depend strongly on evaluation design. 
Joint cropland delineation and crop classification reveal errors hidden by predefined cropland masks, while multi-year evaluation exposes weather-driven distribution shifts, with temperature-based temporal representations improving robustness. 
On SwissCrop25, domain-specific crop mapping models outperform Galileo, with TSViT achieving the highest macro-mIoU and increasing advantages for fine-grained and rare crop classes.
However, no architecture consistently dominates across all operational scenarios.
In-season evaluation reveals a trade-off between architectures: U-TAE performs better early in the season on common crops, whereas TSViT gains an advantage later through improved rare-class discrimination.
These findings demonstrate that operational crop mapping performance cannot be captured by a single benchmark score, but requires evaluation across complementary aspects of deployment. 
SwissCrop25 provides such a benchmark by integrating temporal variability, semantic complexity, and realistic deployment scenarios for systematic comparison of crop mapping approaches.

\newpage
\section*{Acknowledgements}
We thank the anonymous reviewers for their constructive comments. We also thank Manuel Schneider, Chloé Wüst, Sonja Keel (all Agroscope), and Andreas Schellenberger (Federal Office for the Environment) for their valuable input.
This work was supported by the Federal Office for the Environment (FOEN) (06.0091.PZ/0046), the Swiss National Science Foundation (Grant No. 10002727), and the Swiss Federal Office of Agriculture (FOAG) and Agroscope within the Monitoring of the Swiss Agri-Environmental System (MAUS) program. It was enabled by the Swiss Agricultural Landscape Intelligence Platform (SALI) established and maintained by Agroscope. We acknowledge access to Alps at the Swiss National Supercomputing Centre, Switzerland under Agroscope's share with the project ID go57. All funding was awarded to Helge Aasen.
Large language models were used in the preparation of this manuscript for writing assistance, language editing, and code development. All scientific content, experimental results, and conclusions were verified by the authors.

\bibliographystyle{splncs04}
\bibliography{main}

\begin{thebibliography}{10}
\providecommand{\url}[1]{\texttt{#1}}
\providecommand{\urlprefix}{URL }
\providecommand{\doi}[1]{https://doi.org/#1}

\bibitem{asamMappingCropTypes2022}
Asam, S., Gessner, U., Almengor~Gonz{\'a}lez, R., Wenzl, M., Kriese, J., Kuenzer, C.: Mapping {{Crop Types}} of {{Germany}} by {{Combining Temporal Statistical Metrics}} of {{Sentinel-1}} and {{Sentinel-2 Time Series}} with {{LPIS Data}}. Remote Sensing  \textbf{14}(13), ~2981 (Jan 2022). \doi{10.3390/rs14132981}

\bibitem{atzbergerAdvancesRemoteSensing2013}
Atzberger, C.: Advances in {{Remote Sensing}} of {{Agriculture}}: {{Context Description}}, {{Existing Operational Monitoring Systems}} and {{Major Information Needs}}. Remote Sensing  \textbf{5}(2),  949--981 (Feb 2013). \doi{10.3390/rs5020949}

\bibitem{aybarCloudSEN12LargestDataset2024}
Aybar, C., Bautista, L., Montero, D., Contreras, J., Ayala, D., Prudencio, F., Loja, J., Ysuhuaylas, L., Herrera, F., Gonzales, K., Valladares, J., Flores, L.A., Mamani, E., Qui{\~n}onez, M., Fajardo, R., Espinoza, W., Limas, A., Yali, R., Alc{\'a}ntara, A., Leyva, M., {Loayza-Muro}, R., Willems, B., {Mateo-Garc{\'i}a}, G., {G{\'o}mez-Chova}, L.: {{CloudSEN12}}+: {{The}} largest dataset of expert-labeled pixels for cloud and cloud shadow detection in {{Sentinel-2}}. Data in Brief  \textbf{56},  110852 (Oct 2024). \doi{10.1016/j.dib.2024.110852}

\bibitem{barriereBoostingCropClassification2024}
Barriere, V., Claverie, M., Schneider, M., Lemoine, G., {d'Andrimont}, R.: Boosting crop classification by hierarchically fusing satellite, rotational, and contextual data. Remote Sensing of Environment  \textbf{305},  114110 (May 2024). \doi{10.1016/j.rse.2024.114110}

\bibitem{bastonExactextractrFastExtraction2024}
Baston, D.: Exactextractr: {{Fast Extraction}} from {{Raster Datasets}} Using {{Polygons}} (2024). \doi{10.32614/CRAN.package.exactextractr}

\bibitem{becker-reshefCropTypeMaps2023}
{Becker-Reshef}, I., Barker, B., Whitcraft, A., Oliva, P., Mobley, K., Justice, C., Sahajpal, R.: Crop {{Type Maps}} for {{Operational Global Agricultural Monitoring}}. Scientific Data  \textbf{10}(1), ~172 (Mar 2023). \doi{10.1038/s41597-023-02047-9}

\bibitem{boryanMonitoringUSAgriculture2011}
Boryan, C., Yang, Z., Mueller, R., Craig, M.: Monitoring {{US}} agriculture: The {{US Department}} of {{Agriculture}}, {{National Agricultural Statistics Service}}, {{Cropland Data Layer Program}}. Geocarto International  \textbf{26}(5),  341--358 (Aug 2011). \doi{10.1080/10106049.2011.562309}

\bibitem{claverieEuroCropsV20Multiannual2026}
Claverie, M., Chan, A., See, L., Ramos, H., Koeble, R., Yordanov, M., Sk{\o}ien, J.O., Urbano, F., {d'Andrimont}, R., Schneider, M., K{\"o}rner, M., {Van der Velde}, M.: {{EuroCrops}} v2.0: Multi-annual harmonized parcel level crop type data linked to {{European Union-wide}} survey, statistical, and {{Earth Observation}} products. Earth System Science Data  \textbf{18}(6),  4075--4095 (Jun 2026). \doi{10.5194/essd-18-4075-2026}

\bibitem{copernicuslandmonitoringserviceHighResolutionLayer2025}
{Copernicus Land Monitoring Service}: High {{Resolution Layer Croplands}} (2025), \url{https://land.copernicus.eu/en/products/high-resolution-layer-croplands}

\bibitem{cuiClassBalancedLossBased2019a}
Cui, Y., Jia, M., Lin, T.Y., Song, Y., Belongie, S.: Class-{{Balanced Loss Based}} on {{Effective Number}} of {{Samples}}. In: 2019 {{IEEE}}/{{CVF Conference}} on {{Computer Vision}} and {{Pattern Recognition}} ({{CVPR}}). pp. 9260--9269 (Jun 2019). \doi{10.1109/CVPR.2019.00949}

\bibitem{deabelleyraArgentinaNationalMap2025}
{de Abelleyra}, D., Iturralde~Elortegui, M.d.R., Zelaya, K., Portillo, J., Melilli, M., Volante, J., Franzoni, A., L{\'o}pez~Morillo, {\relax CS}., Goyt{\'i}a, Y., Murray, F., Santill{\'a}n, J., Berriolo, J., Lanceta~Pereyra, M., Scavone, A., Continelli, N., Gerlero, G., Salas, D., Reinaldi, J., Lopez~Juane, P., Gomez, D., Krapovickas, S., Sapino, V., Regonat, A., Cracogna, M., Esp{\'i}ndola, C., Valiente, S., Parodi, M., Colombo, F., Scarel, J., Ayala, J., Martins, L., Basanta, M., Rausch, A., Almada, G., Boero, L., Calcha, J., Chiavassa, A., Calandroni, M., Pascal, B., Borracci, S., Erreguerena, J., Besteiro, I., Oyesqui, L., Lazaeta, M., Loizaga, {\relax UD}., Murillo, M., Barrag{\'a}n, M., Ferro, M., Centaure, R., Maekawa, M., Schaber, C., Mart{\'i}n, G., Demateis, F., Varillas, G., Adra, M., Tolosa, E., Coliqueo, M., Kurtz, D., Ybarra, D., Barrios, R., Benedetti, P., Morales, C., Pezzola, A., Winschel, C., Rodriguez~Perez, J., Peralta, A., Ben{\'i}tez, L., German, A., Vitale, {\relax JP}.: Argentina {{National Map}} of {{Crops}} 2024/2025 (Nov 2025). \doi{10.5281/ZENODO.17652711}

\bibitem{druschSentinel2ESAsOptical2012}
Drusch, M., Del~Bello, U., Carlier, S., Colin, O., Fernandez, V., Gascon, F., Hoersch, B., Isola, C., Laberinti, P., Martimort, P., Meygret, A., Spoto, F., Sy, O., Marchese, F., Bargellini, P.: Sentinel-2: {{ESA}}'s {{Optical High-Resolution Mission}} for {{GMES Operational Services}}. Remote Sensing of Environment  \textbf{120},  25--36 (May 2012). \doi{10.1016/j.rse.2011.11.026}

\bibitem{sen4cap2021_validation}
{European Space Agency (ESA)}: {{Sen4CAP}} - {{Sentinels}} for {{Common Agricultural Policy}}: {{Validation Report}}. Tech. rep., European Space Agency (ESA) (2021), \url{https://www.esa-sen4cap.org/wp-content/uploads/files/14_Sen4CAP_VR_v1.2.pdf}

\bibitem{foag2024_kulturflaechen}
{Federal Office for Agriculture (FOAG)}: Landwirtschaftliche {{Kulturfl\"achen}} (2024), \url{https://www.blw.admin.ch/de/landwirtschaftliche-kulturflaechen}

\bibitem{swisstopo_swisstlm3d}
{Federal Office of Topography swisstopo}: {{swissTLM3D}}: {{The}} large-scale topographic landscape model of {{Switzerland}} (2026), \url{https://www.swisstopo.admin.ch/en/landscape-model-swisstlm3d}

\bibitem{fisetteAnnualSpacebasedCrop2014}
Fisette, T., Davidson, A., Daneshfar, B., Rollin, P., Aly, Z., Campbell, L.: Annual space-based crop inventory for {{Canada}}: 2009--2014. In: 2014 {{IEEE Geoscience}} and {{Remote Sensing Symposium}}. pp. 5095--5098 (Jul 2014). \doi{10.1109/IGARSS.2014.6947643}

\bibitem{fisetteAAFCAnnualCrop2013}
Fisette, T., Rollin, P., Aly, Z., Campbell, L., Daneshfar, B., Filyer, P., Smith, A., Davidson, A., Shang, J., Jarvis, I.: {{AAFC}} annual crop inventory. In: 2013 {{Second International Conference}} on {{Agro-Geoinformatics}} ({{Agro-Geoinformatics}}). pp. 270--274 (Aug 2013). \doi{10.1109/Argo-Geoinformatics.2013.6621920}

\bibitem{garioudFLAIRHUBLargescaleMultimodal2026}
Garioud, A., Giordano, S., David, N., Gonthier, N.: {{FLAIR-HUB}}: {{Large-scale}} multimodal dataset for land cover and crop mapping. ISPRS Journal of Photogrammetry and Remote Sensing  \textbf{237},  271--300 (Jul 2026). \doi{10.1016/j.isprsjprs.2026.04.017}

\bibitem{faregarnotPanopticSegmentationSatellite2021}
Garnot, V.S.F., Landrieu, L.: Panoptic {{Segmentation}} of {{Satellite Image Time Series}} with {{Convolutional Temporal Attention Networks}}. In: 2021 {{IEEE}}/{{CVF International Conference}} on {{Computer Vision}} ({{ICCV}}). pp. 4852--4861 (Oct 2021). \doi{10.1109/ICCV48922.2021.00483}

\bibitem{garnotSatelliteImageTime2019}
Garnot, V.S.F., Landrieu, L., Giordano, S., Chehata, N.: Satellite {{Image Time Series Classification}} with {{Pixel-Set Encoders}} and {{Temporal Self-Attention}} (Nov 2019). \doi{10.48550/arXiv.1911.07757}

\bibitem{geodienste_nutzungsflaechen}
{geodienste.ch}: Nutzungsfl\"achen, \url{https://geodienste.ch/services/lwb_nutzungsflaechen}

\bibitem{ghassemiEuropeanUnionCrop2024}
Ghassemi, B., {Izquierdo-Verdiguier}, E., Verhegghen, A., Yordanov, M., Lemoine, G., Moreno~Mart{\'i}nez, {\'A}., De~Marchi, D., {van der Velde}, M., Vuolo, F., {d'Andrimont}, R.: European {{Union}} crop map 2022: {{Earth}} observation's 10-meter dive into {{Europe}}'s crop tapestry. Scientific Data  \textbf{11}(1), ~1048 (Sep 2024). \doi{10.1038/s41597-024-03884-y}

\bibitem{hollandComplyingConservationCompliance2020}
Holland, A., Bennett, D., Secchi, S.: Complying with conservation compliance? {{An}} assessment of recent evidence in the {{US Corn Belt}}. Environmental Research Letters  \textbf{15}(8),  084035 (Aug 2020). \doi{10.1088/1748-9326/ab8f60}

\bibitem{ipcc2006guidelines}
{Intergovernmental Panel on Climate Change (IPCC)}: 2006 {{IPCC Guidelines}} for {{National Greenhouse Gas Inventories}}: {{Volume}} 4: {{Agriculture}}, {{Forestry}} and {{Other Land Use}}. Tech. rep., Institute for Global Environmental Strategies (IGES), Hayama, Kanagawa, Japan (2006), \url{https://www.ipcc-nggip.iges.or.jp/public/2006gl/vol4.html}

\bibitem{kondmannDENETHORDynamicEarthNETDataset2021}
Kondmann, L., Toker, A., Russwurm, M., Camero, A., Peressuti, D., Milcinski, G., Long{\'e}p{\'e}, N., Mathieu, P.P., Davis, T., Marchisio, G., {Leal-Taix{\'e}}, L., Zhu, X.X.: {{DENETHOR}}: {{The DynamicEarthNET}} dataset for {{Harmonized}}, inter-{{Operable}}, analysis-{{Ready}}, daily crop monitoring from space. In: 35th {{Conference}} on {{Neural Information Processing Systems Datasets}} and {{Benchmarks Track}}. pp. 1--13. Virtual (Dec 2021), \url{https://datasets-benchmarks-proceedings.neurips.cc/paper/2021/file/5b8add2a5d98b1a652ea7fd72d942dac-Paper-round2.pdf}

\bibitem{liAutomated10mResolution2026}
Li, H., Di, L., Zhang, C., Guo, L., Yu, E.G., Shao, B., Liu, Z., Li, H.: Automated 10-m {{Resolution In-season Crop-type Data Layer Mapping}} for {{Contiguous United States}}. Scientific Data  \textbf{13}(1), ~750 (Mar 2026). \doi{10.1038/s41597-026-07099-1}

\bibitem{meteoswiss2021_tabsd}
{MeteoSwiss}: Documentation of {{MeteoSwiss Grid-Data Products}}: {{Daily Mean}}, {{Minimum}} and {{Maximum Temperature}}: {{TabsD}}, {{TminD}}, {{TmaxD}}. Tech. rep., {Federal Office of Meteorology and Climatology MeteoSwiss}, Z\"urich, Switzerland (2021), \url{https://www.meteoschweiz.admin.ch/dam/jcr:818a4d17-cb0c-4e8b-92c6-1a1bdf5348b7/ProdDoc_TabsD.pdf}

\bibitem{microsoftopensourceMicrosoftPlanetaryComputerOctober2022}
{Microsoft Open Source}, McFarland, M., Emanuele, R., Morris, D., Augspurger, T.: Microsoft/{{PlanetaryComputer}}: {{October}} 2022. Zenodo (Oct 2022). \doi{10.5281/ZENODO.7261897}

\bibitem{monteroFacilitatingAdvancedSentinel22024}
Montero, D., Mahecha, M.D., Aybar, C., Mosig, C., Wieneke, S.: Facilitating advanced {{Sentinel-2}} analysis through a simplified computation of {{Nadir BRDF Adjusted Reflectance}}. The International Archives of the Photogrammetry, Remote Sensing and Spatial Information Sciences  \textbf{XLVIII-4/W12-2024},  105--112 (Jun 2024). \doi{10.5194/isprs-archives-XLVIII-4-W12-2024-105-2024}

\bibitem{nyborgGeneralizedClassificationSatellite2022}
Nyborg, J., Pelletier, C., Assent, I.: Generalized {{Classification}} of {{Satellite Image Time Series}} with {{Thermal Positional Encoding}} (Jun 2022). \doi{10.48550/arXiv.2203.09175}

\bibitem{nyborgTimeMatchUnsupervisedCrossregion2022}
Nyborg, J., Pelletier, C., Lef{\`e}vre, S., Assent, I.: {{TimeMatch}}: {{Unsupervised}} cross-region adaptation by temporal shift estimation. ISPRS Journal of Photogrammetry and Remote Sensing  \textbf{188},  301--313 (Jun 2022). \doi{10.1016/j.isprsjprs.2022.04.018}

\bibitem{pelletierTemporalConvolutionalNeural2019}
Pelletier, C., Webb, G.I., Petitjean, F.: Temporal {{Convolutional Neural Network}} for the {{Classification}} of {{Satellite Image Time Series}}. Remote Sensing  \textbf{11}(5), ~523 (Jan 2019). \doi{10.3390/rs11050523}

\bibitem{phamTemporallyTransferableCrop2024}
Pham, V.D., Tetteh, G., Thiel, F., Erasmi, S., Schwieder, M., Frantz, D., {van der Linden}, S.: Temporally transferable crop mapping with temporal encoding and deep learning augmentations. International Journal of Applied Earth Observation and Geoinformation  \textbf{129},  103867 (May 2024). \doi{10.1016/j.jag.2024.103867}

\bibitem{reussEuroCropsMLTimeSeries2025}
Reuss, J., Macdonald, J., Becker, S., Richter, L., K{\"o}rner, M.: The {{EuroCropsML}} time series benchmark dataset for few-shot crop type classification in {{Europe}}. Scientific Data  \textbf{12}(1), ~664 (Apr 2025). \doi{10.1038/s41597-025-04952-7}

\bibitem{russwurmMultiTemporalLandCover2018}
Ru{\ss}wurm, M., K{\"o}rner, M.: Multi-{{Temporal Land Cover Classification}} with {{Sequential Recurrent Encoders}}. ISPRS International Journal of Geo-Information  \textbf{7}(4), ~129 (Apr 2018). \doi{10.3390/ijgi7040129}

\bibitem{russwurmSelfattentionRawOptical2020}
Ru{\ss}wurm, M., K{\"o}rner, M.: Self-attention for raw optical {{Satellite Time Series Classification}}. ISPRS Journal of Photogrammetry and Remote Sensing  \textbf{169},  421--435 (Nov 2020). \doi{10.1016/j.isprsjprs.2020.06.006}

\bibitem{russwurmBreizhCropsTimeSeries2019}
Ru{\ss}wurm, M., Pelletier, C., Zollner, M., Lef{\`e}vre, S., K{\"o}rner, M.: {{BreizhCrops}}: {{A Time Series Dataset}} for {{Crop Type Mapping}} (May 2020). \doi{10.48550/arXiv.1905.11893}

\bibitem{schneiderEuroCropsLargestHarmonized2023}
Schneider, M., Schelte, T., Schmitz, F., K{\"o}rner, M.: {{EuroCrops}}: {{The Largest Harmonized Open Crop Dataset Across}} the {{European Union}}. Scientific Data  \textbf{10}(1), ~612 (Sep 2023). \doi{10.1038/s41597-023-02517-0}

\bibitem{senarasEarlySeasonCropClassification2024}
Senaras, C., Holden, P., Davis, T., Wania, A., Rana, A.S., Grady, M., De~Jeu, R.: Early-{{Season Crop Classification}} with {{Planet Fusion}}. In: {{IGARSS}} 2024 - 2024 {{IEEE International Geoscience}} and {{Remote Sensing Symposium}}. pp. 4145--4149 (Jul 2024). \doi{10.1109/IGARSS53475.2024.10642187}

\bibitem{sykasSentinel2MultiyearMulticountry2022}
Sykas, D., Sdraka, M., Zografakis, D., Papoutsis, I.: A {{Sentinel-2 Multiyear}}, {{Multicountry Benchmark Dataset}} for {{Crop Classification}} and {{Segmentation With Deep Learning}}. IEEE Journal of Selected Topics in Applied Earth Observations and Remote Sensing  \textbf{15},  3323--3339 (2022). \doi{10.1109/JSTARS.2022.3164771}

\bibitem{tarasiouViTsSITSVision2023}
Tarasiou, M., Chavez, E., Zafeiriou, S.: {{ViTs}} for {{SITS}}: {{Vision Transformers}} for {{Satellite Image Time Series}} (Apr 2023). \doi{10.48550/arXiv.2301.04944}

\bibitem{tettehAgriculturalLandUse2025}
Tetteh, G.O., Schwieder, M., Blickensd{\"o}rfer, L., Gocht, A., Erasmi, S.: Agricultural land use (raster): {{National-scale}} crop type maps for {{Germany}} from combined time series of {{Sentinel-2}} and {{Landsat}} data (2025) (Sep 2025). \doi{10.5281/zenodo.17181502}

\bibitem{tsengLightweightPretrainedTransformers2024}
Tseng, G., Cartuyvels, R., Zvonkov, I., Purohit, M., Rolnick, D., Kerner, H.: Lightweight, {{Pre-trained Transformers}} for {{Remote Sensing Timeseries}} (Feb 2024). \doi{10.48550/arXiv.2304.14065}

\bibitem{pmlr-v267-tseng25a}
Tseng, G., Fuller, A., Reil, M., Herzog, H., Beukema, P., Bastani, F., Green, J.R., Shelhamer, E., Kerner, H., Rolnick, D.: Galileo: {{Learning Global}} \& {{Local Features}} of {{Many Remote Sensing Modalities}}. In: Singh, A., Fazel, M., Hsu, D., {Lacoste-Julien}, S., Berkenkamp, F., Maharaj, T., Wagstaff, K., Zhu, J. (eds.) Proceedings of the 42nd International Conference on Machine Learning. Proceedings of Machine Learning Research, vol.~267, pp. 60280--60300. PMLR (Jul 2025), \url{https://proceedings.mlr.press/v267/tseng25a.html}

\bibitem{turkogluCropMappingImage2021}
Turkoglu, M.O., D'Aronco, S., Perich, G., Liebisch, F., Streit, C., Schindler, K., Wegner, J.D.: Crop mapping from image time series: {{Deep}} learning with multi-scale label hierarchies. Remote Sensing of Environment  \textbf{264},  112603 (Oct 2021). \doi{10.1016/j.rse.2021.112603}

\bibitem{turkogluT3SThinkThermal2026}
Turkoglu, M.O., Ledain, S., Zweidler, J., Lauber, T., Aasen, H.: {{T3S}}: {{Think}} in {{Thermal Time}} for {{Generalizable Crop Mapping}} from {{Satellite Image Time Series}} (Jul 2026). \doi{10.48550/arXiv.2506.12885}

\bibitem{vantrichtWorldCerealDynamicOpensource2023}
Van~Tricht, K., Degerickx, J., Gilliams, S., Zanaga, D., Battude, M., Grosu, A., Brombacher, J., Lesiv, M., Bayas, J.C.L., Karanam, S., Fritz, S., {Becker-Reshef}, I., Franch, B., {Moll{\`a}-Bononad}, B., Boogaard, H., Pratihast, A.K., Koetz, B., Szantoi, Z.: {{WorldCereal}}: A dynamic open-source system for global-scale, seasonal, and reproducible crop and irrigation mapping. Earth System Science Data  \textbf{15}(12),  5491--5515 (Dec 2023). \doi{10.5194/essd-15-5491-2023}

\bibitem{vincentPixelwiseAgriculturalImage2024}
Vincent, E., Ponce, J., Aubry, M.: Pixel-wise {{Agricultural Image Time Series Classification}}: {{Comparisons}} and a {{Deformable Prototype-based Approach}} (Jul 2024). \doi{10.48550/arXiv.2303.12533}

\bibitem{weikmannTimeSen2CropMillionLabeled2021}
Weikmann, G., Paris, C., Bruzzone, L.: {{TimeSen2Crop}}: {{A Million Labeled Samples Dataset}} of {{Sentinel}} 2 {{Image Time Series}} for {{Crop-Type Classification}}. IEEE Journal of Selected Topics in Applied Earth Observations and Remote Sensing  \textbf{14},  4699--4708 (Apr 2021). \doi{10.1109/JSTARS.2021.3073965}

\bibitem{wijesinghaEvaluatingSpatialTemporal2024}
Wijesingha, J., Dzene, I., Wachendorf, M.: Evaluating the spatial--temporal transferability of models for agricultural land cover mapping using {{Landsat}} archive. ISPRS Journal of Photogrammetry and Remote Sensing  \textbf{213},  72--86 (Jul 2024). \doi{10.1016/j.isprsjprs.2024.05.020}

\bibitem{wuChallengesOpportunitiesRemote2023}
Wu, B., Zhang, M., Zeng, H., Tian, F., Potgieter, A.B., Qin, X., Yan, N., Chang, S., Zhao, Y., Dong, Q., Boken, V., Plotnikov, D., Guo, H., Wu, F., Zhao, H., Deronde, B., Tits, L., Loupian, E.: Challenges and opportunities in remote sensing-based crop monitoring: A review. National Science Review  \textbf{10}(4),  nwac290 (Apr 2023). \doi{10.1093/nsr/nwac290}

\bibitem{yuanEmpiricalStudyData2025}
Yuan, Y., Lin, L., Xin, Q., Zhou, Z.G., Liu, Q.: An {{Empirical Study}} on {{Data Augmentation}} for {{Pixelwise Satellite Image Time-Series Classification}} and {{Cross-Year Adaptation}}. IEEE Journal of Selected Topics in Applied Earth Observations and Remote Sensing  \textbf{18},  5172--5188 (2025). \doi{10.1109/JSTARS.2025.3527017}

\end{thebibliography}

\clearpage
\setcounter{section}{0}
\renewcommand{\thesection}{\Alph{section}}
\renewcommand{\theHsection}{supp.sec.\arabic{section}}
\renewcommand{\theHtable}{supp.tab.\arabic{table}}
\renewcommand{\theHfigure}{supp.fig.\arabic{figure}}
\setcounter{table}{0}
\setcounter{figure}{0}

\section*{Supplementary Material}
\addcontentsline{toc}{section}{Supplementary Material}

This supplementary provides preprocessing details, evaluation metric definitions, training hyperparameters, and extended benchmark results mirroring the experiment order of the main paper (scene completeness, temporal generalisation, fine-grained classification, in-season usability, efficiency), followed by per-class results for all \nrCropClasses\ agricultural crop classes.


\section{Label Preprocessing}
\label{supp:preprocessing}

\paragraph{LNF--swissTLM3D merge.}
LNF parcel polygons and swissTLM3D land cover polygons are combined into a single label layer per year using a priority-based overlap resolution. This ordering preserves LNF parcel labels for regular agricultural fields. Alpine summer pastures (LNF code~930, \textit{Sömmerungsweiden}) are refined using swissTLM3D forest and unproductive terrain masks, and administrative LNF code~998 is assigned the lowest priority.

\paragraph{Road and railway polygonisation.}
SwissTLM3D encodes roads and railways as line geometries. For rasterisation, these are converted to polygons by buffering each feature by half its class-specific nominal width: paths (\SI{1}{\meter}), tracks (\SI{2}{\meter}), minor roads (\SI{3}{\meter}--\SI{4}{\meter}), main roads (\SI{6}{\meter}--\SI{10}{\meter}), motorways (\SI{30}{\meter}), railways (\SI{8}{\meter} single-track, \SI{13}{\meter} double-track). Underground structures (tunnels, underpasses) are excluded.

\paragraph{Geometry cleaning.}
All vector geometries are validated with Shapely's \texttt{make\_valid}, snapped to a \SI{1}{\meter} coordinate grid to eliminate floating-point precision artefacts, and filtered to retain only polygon and multipolygon types.

\paragraph{Rasterisation.}
The merged polygon layer is rasterised at \SI{10}{\meter} resolution to produce pixel-level labels for the Sentinel-2 data cubes.
Rather than the centre-of-pixel rule (as in \texttt{gdal\_rasterize}), we use a fractional coverage approach: for each pixel, we compute the area fraction covered by each of the \nrLNFCodes{} source classes using the \texttt{coverage\_fraction} function from the \texttt{exactextractr} package~\cite{bastonExactextractrFastExtraction2024}, aggregate these fractions to the \nrClasses{} modelled classes, and assign the majority class.
    Aggregating before taking the majority ensures that fractions belonging to the same modelled class are pooled first, rather than competing against each other.
This area-weighted assignment is more accurate than centre-of-pixel for narrow field strips and small parcels, and mirrors how a satellite sensor integrates over its footprint.

\newpage
\section{Per-Year Dataset Composition}
\label{supp:dataset_stats}

\Cref{supp:tab:dataset_stats} reports the per-year breakdown of parcels, cubes, and land area. Partial years exclude cantons with insufficient LNF coverage.

\begin{table}[H]
\centering
\caption{Per-year dataset composition. Ag: agricultural area; Non-crop: swissTLM3D land cover area. Partial years exclude cantons with $<$90\% LNF coverage.}
\label{supp:tab:dataset_stats}
\setlength{\tabcolsep}{5pt}
\begin{tabular}{lrrrrl}
\toprule
\textbf{Year} & \textbf{Parcels} & \textbf{Cubes} & \textbf{Ag (kha)} & \textbf{Non-crop (kha)} & \textbf{Coverage} \\
\midrule
2019 &    740{,}597 & 10{,}546 &   828 & 2{,}307 & Partial \\
2020 & 1{,}536{,}321 & 21{,}439 & 1{,}022 & 2{,}294 & Partial \\
2021 & 2{,}023{,}603 & 26{,}089 & 1{,}268 & 2{,}248 & Complete \\
2022 & 2{,}004{,}326 & 26{,}089 & 1{,}245 & 2{,}252 & Complete \\
2023 & 2{,}074{,}936 & 26{,}089 & 1{,}266 & 2{,}256 & Complete \\
2024 & 2{,}116{,}874 & 26{,}090 & 1{,}284 & 2{,}264 & Complete \\
2025 & 2{,}146{,}317 & 26{,}843 & 1{,}306 & 2{,}270 & Complete \\
\bottomrule
\end{tabular}
\end{table}


\clearpage
\section{Class Distribution}
\label{supp:classdist}

\Cref{fig:label_dist} shows the full class distribution of SwissCrop25 across all \nrClasses\ classes.

\noindent\begin{minipage}{\linewidth}
  \centering
  \includegraphics[width=\linewidth,height=\textheight,keepaspectratio]{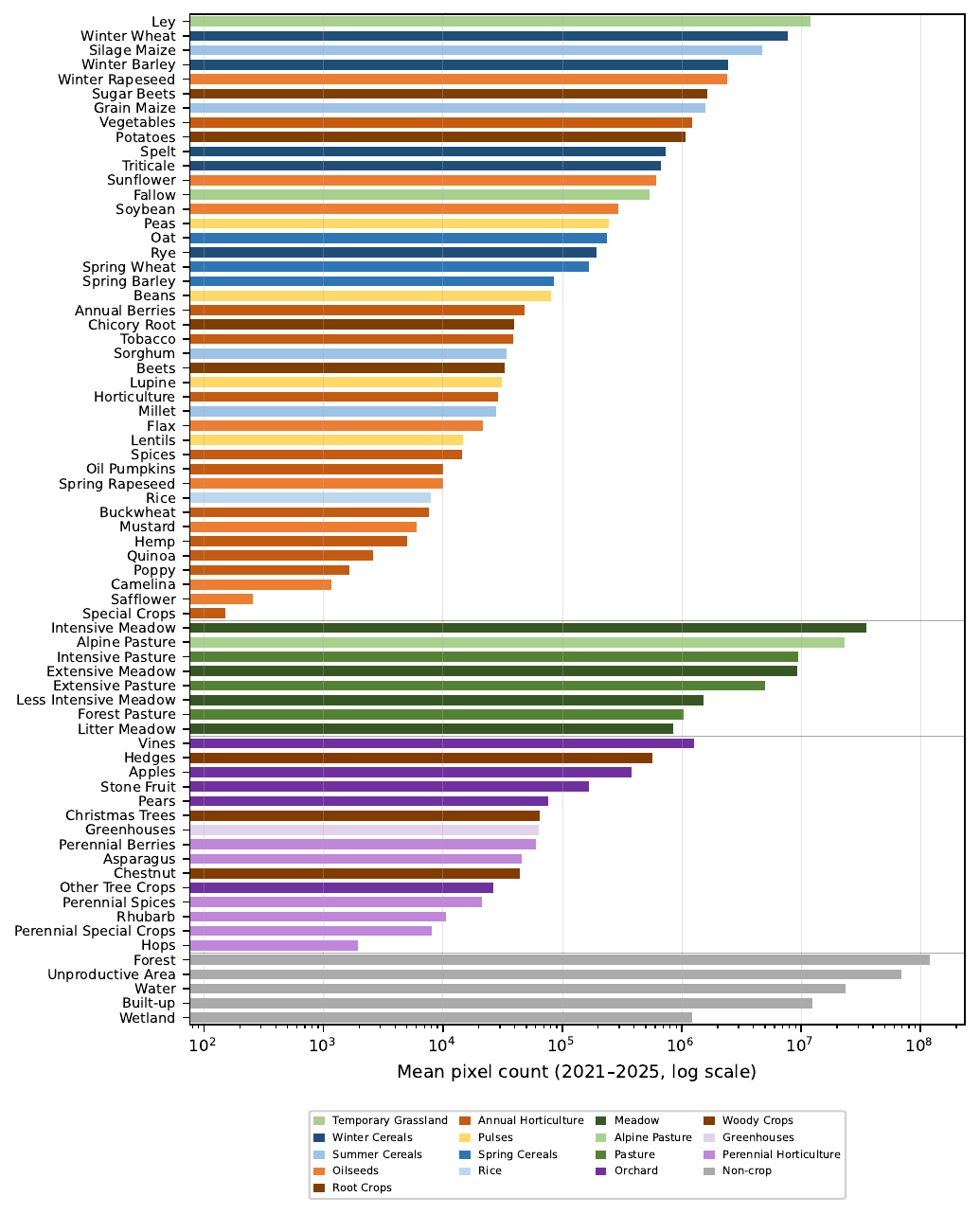}
  \captionof{figure}{Class distribution of SwissCrop25 across all \nrClasses\ classes (\nrCropClasses\ agricultural and \nrNonCropClasses\ non-crop land cover), measured as mean pixel-equivalent area averaged over the five complete years (2021--2025). Classes are sorted by frequency within each taxonomy group (Arable Land, Grassland, Permanent, Non-crop); colours indicate lv2 subgroup. The distribution spans five orders of magnitude, with class imbalance exceeding 200{,}000:1.}
  \label{fig:label_dist}
\end{minipage}


\section{Evaluation Metrics}
\label{supp:metrics}

\paragraph{Classification metrics.}
\textbf{OA} (Overall Accuracy) measures the fraction of correctly classified pixels across the \nrCropClasses\ agricultural classes.
\textbf{GIoU} (Global IoU) computes intersection-over-union globally over all agricultural pixels and is equivalent to a frequency-weighted mean IoU.
\textbf{mIoU} (mean IoU) and \textbf{mF1} (macro-F1) are class-balanced metrics obtained by averaging per-class IoU and F1 scores, respectively, across the \nrCropClasses\ agricultural classes.

\paragraph{Calibration metrics.}
\textbf{ECE} (Expected Calibration Error) bins predictions into 15 equal-width confidence bins and reports the weighted mean absolute deviation between mean confidence and accuracy within each bin.
\textbf{NLL} (Negative Log-Likelihood) is the mean per-pixel negative log-likelihood under the softmax output distribution, computed over all labelled pixels without class weighting.


\section{Training Details}
\label{supp:training}

All models are trained with AdamW using a peak learning rate of $10^{-3}$, weight decay $0.01$, and a cosine decay schedule with 5\% linear warmup.
Training runs for 15 epochs on 4 NVIDIA GH200 GPUs with an effective batch size of 64 (achieved via gradient accumulation where needed).
For Galileo-nano, the encoder is fine-tuned at $10^{-4}$ (0.1$\times$ the head learning rate).
All runs use class-balanced cross-entropy loss~\cite{cuiClassBalancedLossBased2019a} ($\beta=0.99999$) and a fixed seed of 7777.

\begin{table}[H]
\centering
\caption{Per-model training hyperparameters. Effective batch size = batch size $\times$ GPUs $\times$ gradient accumulation steps.}
\label{supp:tab:hyperparams}
\setlength{\tabcolsep}{5pt}
\begin{tabular}{lrrrrr}
\toprule
\textbf{Model} & \textbf{Batch} & \textbf{Accum.} & \textbf{Eff.\ batch} & \textbf{Head LR} & \textbf{Enc.\ LR} \\
\midrule
U-TAE          & 16 & 1 & 64 & $10^{-3}$ & --- \\
TSViT          &  4 & 4 & 64 & $10^{-3}$ & --- \\
Galileo-nano   &  4 & 4 & 64 & $10^{-3}$ & $10^{-4}$ \\
\bottomrule
\end{tabular}
\end{table}

\paragraph{Low-resource protocol.}
For the low-resource evaluation (\cref{supp:tab:lowresource}), each model is trained on a randomly sampled 10\% subset of the training cubes for that split.
The subset is drawn once with a fixed seed and held constant across all models to ensure a fair comparison.
Evaluation uses the identical protocol as the full-data setting.

\newpage

\section{Scene Completeness}
\label{supp:extended}

\Cref{supp:tab:mask_full} extends \cref{tab:mask} of the main paper with Precision, Recall, and F1, and includes frozen encoder variants.

\begin{table}[H]
\centering
\caption{Full binary agricultural mask evaluation including frozen encoder variants. Metrics averaged over all five LOYO splits. Best per column is \textbf{bold}.}
\label{supp:tab:mask_full}
\resizebox{\textwidth}{!}{%
\begin{tabular}{lrrrr}
\toprule
\textbf{Model} & \textbf{Precision (\%) $\uparrow$} & \textbf{Recall (\%) $\uparrow$} & \textbf{F1 (\%) $\uparrow$} & \textbf{IoU$_\mathrm{ag}$ (\%) $\uparrow$} \\
\midrule
U-TAE & 97.2 & \textbf{91.6} & \textbf{94.3} & \textbf{89.2} \\
TSViT & \textbf{97.4} & 91.0 & 94.1 & 88.9 \\
Galileo-nano & 96.4 & 89.0 & 92.5 & 86.1 \\
Galileo-nano (frozen) & 95.3 & 85.3 & 90.0 & 81.8 \\
Galileo-base (frozen) & 95.7 & 87.3 & 91.3 & 83.9\\
\bottomrule
\end{tabular}
}
\end{table}

To further contextualise the benchmark results, we evaluate two naive temporal baselines that exploit temporal label persistence. Rather than learning from satellite imagery, both directly reuse ground-truth annotations from other years and therefore represent reference points rather than operational methods. 
The \textbf{previous-year} baseline assigns each pixel the label from the preceding year's annotation.
The \textbf{majority-vote} baseline assigns the most frequent label across the three remaining years.
Both achieve $>$99\% IoU$_\mathrm{ag}$ on the binary agricultural mask evaluation, reflecting the high year-to-year spatial stability of the Swiss agricultural landscape within the LOYO splits. 
However, this stability cannot be assumed over longer time horizons or in regions without comparable annual agricultural registries.
Crop type classification results with the naive temporal baselines are reported in \cref{supp:baselines}.


\newpage
\section{Temporal Generalisation}
\label{supp:temporal}

\subsection*{Full baseline results}
\Cref{supp:tab:results_full} extends \cref{tab:results} of the main paper with frozen encoder variants.

\begin{table}[H]
\centering
\caption{Full baseline results including frozen encoder variants, under the LOYO protocol. See \cref{tab:results} in the main paper for metric definitions. Best value per split and metric is \textbf{bold}.}
\label{supp:tab:results_full}
\setlength{\tabcolsep}{4pt}
\resizebox{\linewidth}{!}{
\begin{tabular}{llrrrrrr}
\toprule
\textbf{Test year} & \textbf{Model} & \textbf{OA (\%) $\uparrow$} & \textbf{GIoU (\%) $\uparrow$} & \textbf{mIoU (\%) $\uparrow$} & \textbf{mF1 (\%) $\uparrow$} & \textbf{ECE (\%) $\downarrow$} & \textbf{NLL\,(-) $\downarrow$} \\
\midrule
2021 & U-TAE & \textbf{77.4} & \textbf{63.1} & 36.5 & 46.5 & \textbf{0.99} & \textbf{0.39} \\
 & TSViT & 76.9 & 62.5 & \textbf{46.6} & \textbf{58.9} & 2.36 & 0.41 \\
 & Galileo-nano (FT) & 73.0 & 57.5 & 30.5 & 41.3 & 1.06 & 0.50 \\
 & Galileo-nano (frozen) & 61.6 & 44.5 & 12.7 & 18.1 & 1.09 & 0.68 \\
 & Galileo-base (frozen) & 65.8 & 49.1 & 19.4 & 27.8 & 1.42 & 0.60 \\
\midrule
2022 & U-TAE & \textbf{79.2} & \textbf{65.5} & 37.6 & 47.4 & 0.12 & \textbf{0.36} \\
 & TSViT & 78.7 & 64.9 & \textbf{50.0} & \textbf{62.8} & 0.13 & \textbf{0.36} \\
 & Galileo-nano (FT) & 73.8 & 58.5 & 31.3 & 42.2 & 0.55 & 0.48 \\
 & Galileo-nano (frozen) & 64.2 & 47.3 & 15.1 & 21.1 & 0.22 & 0.63 \\
 & Galileo-base (frozen) & 68.0 & 51.5 & 21.1 & 29.5 & \textbf{0.11} & 0.55 \\
\midrule
2023 & U-TAE & \textbf{78.2} & \textbf{64.2} & 33.7 & 43.1 & 0.35 & \textbf{0.38} \\
 & TSViT & 77.0 & 62.6 & \textbf{47.9} & \textbf{60.6} & 1.71 & 0.40 \\
 & Galileo-nano (FT) & 72.2 & 56.5 & 28.3 & 38.6 & 0.65 & 0.50 \\
 & Galileo-nano (frozen) & 62.1 & 45.0 & 14.3 & 20.3 & \textbf{0.15} & 0.67 \\
 & Galileo-base (frozen) & 66.4 & 49.7 & 20.0 & 28.1 & 0.70 & 0.58 \\
\midrule
2024 & U-TAE & \textbf{75.3} & \textbf{60.4} & 33.2 & 43.2 & 1.09 & \textbf{0.44} \\
 & TSViT & \textbf{75.3} & \textbf{60.4} & \textbf{44.6} & \textbf{57.7} & 2.48 & 0.45 \\
 & Galileo-nano (FT) & 71.0 & 55.1 & 28.3 & 38.9 & \textbf{1.00} & 0.54 \\
 & Galileo-nano (frozen) & 61.5 & 44.4 & 12.8 & 18.4 & 1.62 & 0.70 \\
 & Galileo-base (frozen) & 66.0 & 49.3 & 19.4 & 27.6 & 1.74 & 0.61 \\
\midrule
2025 & U-TAE & \textbf{78.4} & \textbf{64.4} & 38.3 & 48.3 & 1.29 & \textbf{0.40} \\
 & TSViT & 77.6 & 63.4 & \textbf{51.3} & \textbf{63.6} & 2.60 & 0.42 \\
 & Galileo-nano (FT) & 74.4 & 59.2 & 33.6 & 44.6 & \textbf{0.34} & 0.49 \\
 & Galileo-nano (frozen) & 64.0 & 47.1 & 15.7 & 21.9 & 0.98 & 0.64 \\
 & Galileo-base (frozen) & 69.2 & 52.9 & 22.8 & 31.7 & 0.81 & 0.55 \\
\midrule
\midrule
Mean $\pm$ Std & U-TAE & \textbf{77.7} $\pm$ 1.5 & \textbf{63.5} $\pm$ 2.0 & 35.8 $\pm$ 2.3 & 45.7 $\pm$ 2.4 & 0.77 $\pm$ 0.50 & \textbf{0.40} $\pm$ 0.03 \\
 & TSViT & 77.1 $\pm$ 1.2 & 62.7 $\pm$ 1.6 & \textbf{48.1} $\pm$ 2.7 & \textbf{60.7} $\pm$ 2.5 & 1.86 $\pm$ 1.03 & 0.41 $\pm$ 0.03 \\
 & Galileo-nano (FT) & 72.9 $\pm$ 1.3 & 57.4 $\pm$ 1.6 & 30.4 $\pm$ 2.2 & 41.1 $\pm$ 2.5 & \textbf{0.72} $\pm$ 0.31 & 0.50 $\pm$ 0.02 \\
 & Galileo-nano (frozen) & 62.7 $\pm$ 1.3 & 45.6 $\pm$ 1.4 & 14.1 $\pm$ 1.3 & 19.9 $\pm$ 1.6 & 0.81 $\pm$ 0.62 & 0.66 $\pm$ 0.03 \\
 & Galileo-base (frozen) & 67.1 $\pm$ 1.5 & 50.5 $\pm$ 1.7 & 20.6 $\pm$ 1.4 & 28.9 $\pm$ 1.7 & 0.95 $\pm$ 0.64 & 0.58 $\pm$ 0.03\\
\bottomrule
\end{tabular}
}
\end{table}

\newpage
\subsection*{Temporal encoding ablation}
\Cref{supp:tab:temporal_encoding} reports the full per-split results for all temporal encoding variants discussed in \cref{sec:exp:temporal} of the main paper.

\captionof{table}{Per test year, model, and temporal encoding method.
  \textbf{Baseline}: 24 least-cloudy images, uniform calendar spacing (no climate adaptation).
  \textbf{$\boldsymbol{T^{3}S}$}: observations re-indexed to 24 equal GDD bins~\cite{turkogluT3SThinkThermal2026}.
  \textbf{$\boldsymbol{T^{3}S}$\,+\,TPE}: $T^{3}S$ with sinusoidal thermal positional encoding~\cite{nyborgGeneralizedClassificationSatellite2022}.
  Galileo-nano uses a fixed pretrained month-based PE; $T^{3}S$\,+\,TPE is therefore not evaluated.
  Metric definitions in Supp.~\cref{supp:metrics}.}
\label{supp:tab:temporal_encoding}
\fontsize{6.7}{7.7}\selectfont  
\setlength{\tabcolsep}{2pt}     
\begin{longtable}{lllrrrrr}
\toprule
\textbf{Year} & \textbf{Model} & \textbf{Method} & \textbf{OA (\%) $\uparrow$} & \textbf{mIoU (\%) $\uparrow$} & \textbf{mF1 (\%) $\uparrow$} & \textbf{ECE (\%) $\downarrow$} & \textbf{NLL\,(-) $\downarrow$} \\
\midrule\endfirsthead
\multicolumn{8}{l}{\itshape (continued from previous page)}\\[2pt]
\toprule
\textbf{Year} & \textbf{Model} & \textbf{Method} & \textbf{OA (\%) $\uparrow$} & \textbf{mIoU (\%) $\uparrow$} & \textbf{mF1 (\%) $\uparrow$} & \textbf{ECE (\%) $\downarrow$} & \textbf{NLL\,(-) $\downarrow$} \\
\midrule\endhead
\bottomrule\endfoot
2021 & U-TAE & Baseline & 74.1 & 33.3 & 43.0 & 2.26 & 0.48 \\
 &  & $T^{3}S$ & 76.6 & 35.2 & 45.0 & 1.09 & 0.41 \\
 &  & $T^{3}S$ + TPE & 77.4 & 36.5 & 46.5 & 0.99 & 0.39 \\
\cmidrule(l){2-8}
 & TSViT & Baseline & 71.8 & 33.6 & 45.3 & 2.46 & 0.50 \\
 &  & $T^{3}S$ & 73.1 & 35.4 & 47.3 & 1.88 & 0.46 \\
 &  & $T^{3}S$ + TPE & 76.9 & 46.6 & 58.9 & 2.36 & 0.41 \\
\cmidrule(l){2-8}
 & Galileo-nano (FT) & Baseline & 70.8 & 27.1 & 37.6 & 1.14 & 0.50 \\
 &  & $T^{3}S$ & 73.0 & 30.5 & 41.3 & 1.06 & 0.50 \\
\midrule
2022 & U-TAE & Baseline & 78.4 & 36.6 & 46.0 & 0.34 & 0.38 \\
 &  & $T^{3}S$ & 78.3 & 36.3 & 45.5 & 0.47 & 0.38 \\
 &  & $T^{3}S$ + TPE & 79.2 & 37.6 & 47.4 & 0.12 & 0.36 \\
\cmidrule(l){2-8}
 & TSViT & Baseline & 75.1 & 39.3 & 51.6 & 1.11 & 0.42 \\
 &  & $T^{3}S$ & 74.8 & 39.9 & 52.3 & 1.12 & 0.42 \\
 &  & $T^{3}S$ + TPE & 78.7 & 50.0 & 62.8 & 0.13 & 0.36 \\
\cmidrule(l){2-8}
 & Galileo-nano (FT) & Baseline & 73.6 & 30.9 & 41.8 & 0.56 & 0.46 \\
 &  & $T^{3}S$ & 73.8 & 31.3 & 42.2 & 0.55 & 0.48 \\
\midrule
2023 & U-TAE & Baseline & 74.3 & 33.4 & 42.9 & 2.97 & 0.52 \\
 &  & $T^{3}S$ & 76.1 & 34.3 & 44.3 & 1.84 & 0.44 \\
 &  & $T^{3}S$ + TPE & 78.2 & 33.7 & 43.1 & 0.35 & 0.38 \\
\cmidrule(l){2-8}
 & TSViT & Baseline & 73.5 & 35.4 & 47.4 & 1.82 & 0.46 \\
 &  & $T^{3}S$ & 73.3 & 36.2 & 48.2 & 1.94 & 0.46 \\
 &  & $T^{3}S$ + TPE & 77.0 & 47.9 & 60.6 & 1.71 & 0.40 \\
\cmidrule(l){2-8}
 & Galileo-nano (FT) & Baseline & 71.7 & 27.8 & 38.1 & 0.99 & 0.51 \\
 &  & $T^{3}S$ & 72.2 & 28.3 & 38.6 & 0.65 & 0.50 \\
\midrule
2024 & U-TAE & Baseline & 73.0 & 32.7 & 42.4 & 3.31 & 0.55 \\
 &  & $T^{3}S$ & 73.7 & 33.0 & 42.8 & 2.41 & 0.51 \\
 &  & $T^{3}S$ + TPE & 75.3 & 33.2 & 43.2 & 1.09 & 0.44 \\
\cmidrule(l){2-8}
 & TSViT & Baseline & 68.3 & 31.0 & 42.6 & 3.65 & 0.57 \\
 &  & $T^{3}S$ & 71.2 & 33.5 & 45.5 & 2.13 & 0.50 \\
 &  & $T^{3}S$ + TPE & 75.3 & 44.6 & 57.7 & 2.48 & 0.45 \\
\cmidrule(l){2-8}
 & Galileo-nano (FT) & Baseline & 69.7 & 26.7 & 36.9 & 2.45 & 0.54 \\
 &  & $T^{3}S$ & 71.0 & 28.3 & 38.9 & 1.00 & 0.54 \\
\midrule
2025 & U-TAE & Baseline & 78.1 & 37.4 & 46.9 & 1.55 & 0.41 \\
 &  & $T^{3}S$ & 77.5 & 34.1 & 42.4 & 2.01 & 0.43 \\
 &  & $T^{3}S$ + TPE & 78.4 & 38.3 & 48.3 & 1.29 & 0.40 \\
\cmidrule(l){2-8}
 & TSViT & Baseline & 75.3 & 40.5 & 52.5 & 1.85 & 0.44 \\
 &  & $T^{3}S$ & 75.7 & 41.9 & 54.1 & 1.58 & 0.43 \\
 &  & $T^{3}S$ + TPE & 77.6 & 51.3 & 63.6 & 2.60 & 0.42 \\
\cmidrule(l){2-8}
 & Galileo-nano (FT) & Baseline & 73.8 & 31.6 & 42.4 & 1.27 & 0.47 \\
 &  & $T^{3}S$ & 74.4 & 33.6 & 44.6 & 0.34 & 0.49 \\
\midrule
\midrule
Mean & U-TAE & Baseline & 75.6 & 34.7 & 44.2 & 2.09 & 0.47 \\
 &  & $T^{3}S$ & 76.4 & 34.6 & 44.0 & 1.57 & 0.43 \\
 &  & $T^{3}S$ + TPE & 77.7 & 35.8 & 45.7 & 0.77 & 0.40 \\
\cmidrule(l){2-8}
 & TSViT & Baseline & 72.8 & 36.0 & 47.9 & 2.18 & 0.48 \\
 &  & $T^{3}S$ & 73.6 & 37.4 & 49.5 & 1.73 & 0.45 \\
 &  & $T^{3}S$ + TPE & 77.1 & 48.1 & 60.7 & 1.86 & 0.41 \\
\cmidrule(l){2-8}
 & Galileo-nano (FT) & Baseline & 71.9 & 28.8 & 39.4 & 1.28 & 0.50 \\
 &  & $T^{3}S$ & 72.9 & 30.4 & 41.1 & 0.72 & 0.50
\end{longtable}
\normalsize\setlength{\tabcolsep}{6pt}

\newpage
\subsection*{Naive temporal baselines}
\label{supp:baselines}

\Cref{supp:tab:baselines} reports crop type metrics for test years 2022--2025, including naive temporal baselines (defined in \Cref{supp:extended}); the 2021 split is excluded as no prior-year annotations are available for 2020.
TSViT achieves the highest mIoU and mF1 by handling rare arable and permanent classes more effectively, while U-TAE leads on OA, reflecting stronger performance on dominant large-area classes. 
The naive temporal baselines reveal the limits of exploiting temporal persistence, as they achieve near-perfect recall on permanent and grassland classes, but almost completely fail on rotational arable crops (see \cref{supp:fig:confmat_prev_year,supp:fig:confmat_majority}).

\begin{table}[H]
\centering
\caption{Crop type classification metrics for test years 2022--2025, including naive temporal baselines. Metrics are restricted to agricultural classes; ECE and NLL are not reported for baselines as they produce no probabilistic output. Best per column within each split block is \textbf{bold}.}
\label{supp:tab:baselines}
\resizebox{\textwidth}{!}{%
\begin{tabular}{llrrrr}
\toprule
\textbf{Year} & \textbf{Model} & \textbf{OA (\%) $\uparrow$} & \textbf{GIoU (\%) $\uparrow$} & \textbf{mIoU (\%) $\uparrow$} & \textbf{mF1 (\%) $\uparrow$} \\
\midrule
2022 & U-TAE & \textbf{79.2} & \textbf{65.5} & 37.6 & 47.4 \\
 & TSViT & 78.7 & 64.9 & \textbf{50.0} & \textbf{62.8} \\
 & Galileo-nano (FT) & 73.8 & 58.5 & 31.3 & 42.2 \\
 & Previous-year & 75.5 & 60.6 & 37.6 & 41.4 \\
 & Majority vote & 75.2 & 60.2 & 33.9 & 38.8 \\
\midrule
2023 & U-TAE & \textbf{78.2} & \textbf{64.2} & 33.7 & 43.1 \\
 & TSViT & 77.0 & 62.6 & \textbf{47.9} & \textbf{60.6} \\
 & Galileo-nano (FT) & 72.2 & 56.5 & 28.3 & 38.6 \\
 & Previous-year & 75.0 & 60.0 & 37.0 & 41.1 \\
 & Majority vote & 77.0 & 62.7 & 37.2 & 41.1 \\
\midrule
2024 & U-TAE & 75.3 & 60.4 & 33.2 & 43.2 \\
 & TSViT & 75.3 & 60.4 & \textbf{44.6} & \textbf{57.7} \\
 & Galileo-nano (FT) & 71.0 & 55.1 & 28.3 & 38.9 \\
 & Previous-year & 75.4 & 60.6 & 37.2 & 41.2 \\
 & Majority vote & \textbf{76.8} & \textbf{62.3} & 38.0 & 41.9 \\
\midrule
2025 & U-TAE & \textbf{78.4} & \textbf{64.4} & 38.3 & 48.3 \\
 & TSViT & 77.6 & 63.4 & \textbf{51.3} & \textbf{63.6} \\
 & Galileo-nano (FT) & 74.4 & 59.2 & 33.6 & 44.6 \\
 & Previous-year & 76.0 & 61.3 & 38.5 & 42.2 \\
 & Majority vote & 76.0 & 61.3 & 35.7 & 40.3 \\
\midrule
\midrule
Mean $\pm$ Std & U-TAE & \textbf{77.8} $\pm$ 1.7 & \textbf{63.6} $\pm$ 2.2 & 35.7 $\pm$ 2.6 & 45.5 $\pm$ 2.7 \\
 & TSViT & 77.1 $\pm$ 1.4 & 62.8 $\pm$ 1.9 & \textbf{48.4} $\pm$ 2.9 & \textbf{61.2} $\pm$ 2.6 \\
 & Galileo-nano (FT) & 72.9 $\pm$ 1.5 & 57.3 $\pm$ 1.9 & 30.4 $\pm$ 2.6 & 41.1 $\pm$ 2.9 \\
 & Previous-year & 75.5 $\pm$ 0.4 & 60.6 $\pm$ 0.5 & 37.6 $\pm$ 0.7 & 41.4 $\pm$ 0.5 \\
 & Majority vote & 76.3 $\pm$ 0.8 & 61.6 $\pm$ 1.1 & 36.2 $\pm$ 1.8 & 40.5 $\pm$ 1.3\\
\bottomrule
\end{tabular}
}
\end{table}

\subsection*{Winter cereal confusions (2024)}

\begin{figure}[H]
  \centering
  \includegraphics[width=\linewidth]{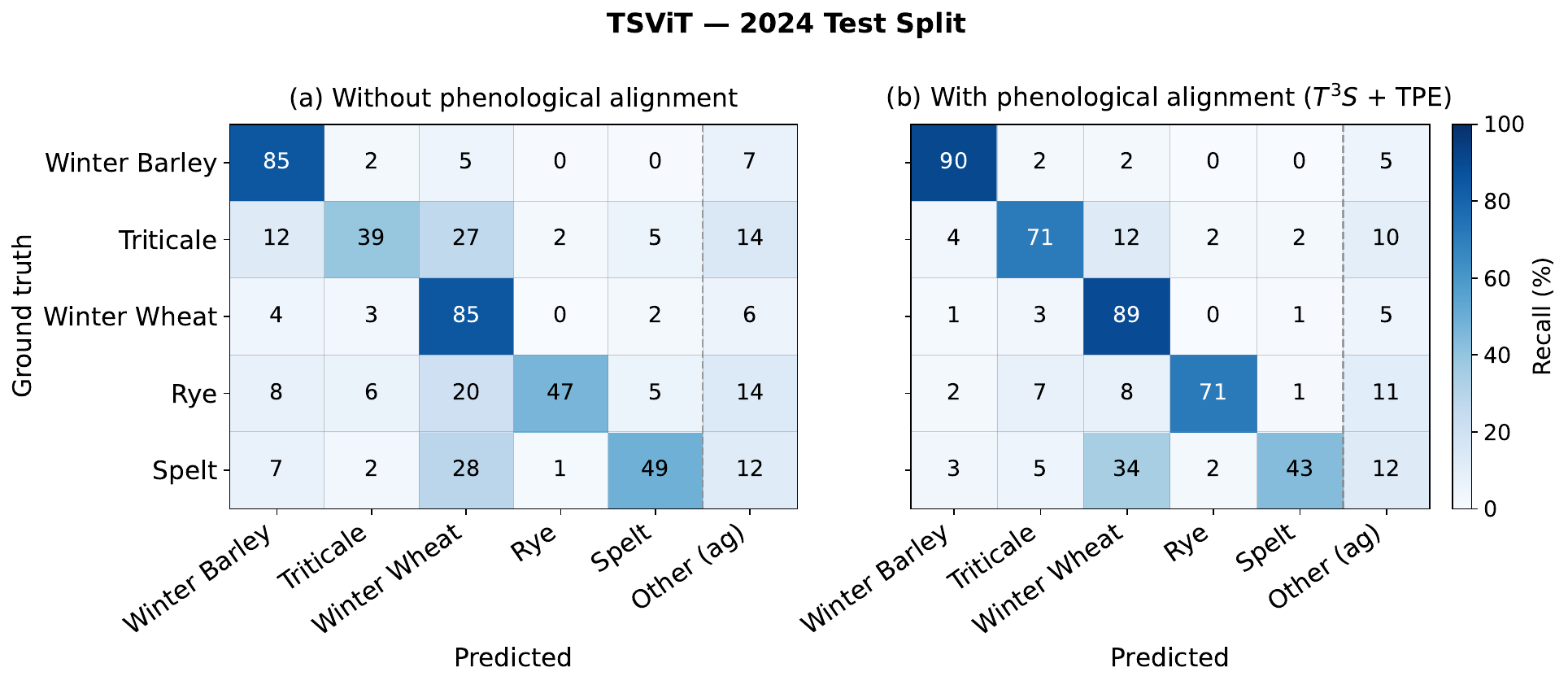}
  \caption{Row-normalised confusion matrix (recall, \%) for the five winter cereal classes, TSViT 2024 test split.
    \textbf{(a)} Without phenological alignment (DOY sampling with cloud filtering).
    \textbf{(b)} With phenological alignment ($T^{3}S$\,+\,TPE: GDD-bin reindexing and thermal positional encoding).
    The \emph{Other} column (grey, dashed separator) aggregates all predictions outside the winter cereal group.
    In the baseline, Triticale and Rye are frequently confused with Winter Wheat; $T^{3}S$\,+\,TPE reduces these confusions and recovers both classes to above 70\% recall.
    Spelt shows little benefit from thermal alignment, with recall remaining largely unchanged.
    The 2024 phenological anomaly visible in \cref{fig:gdd_ndvi} thus manifests as systematic within-group confusion among minority winter cereals, partially recovered by explicit thermal encoding.}
  \label{supp:fig:confmat_winter_cereals_2024}
\end{figure}


\newpage
\section{Fine-Grained Classification}
\label{supp:classification}

\subsection*{Full confusion matrices}

\begin{figure}[H]
  \centering
  \includegraphics[width=\linewidth]{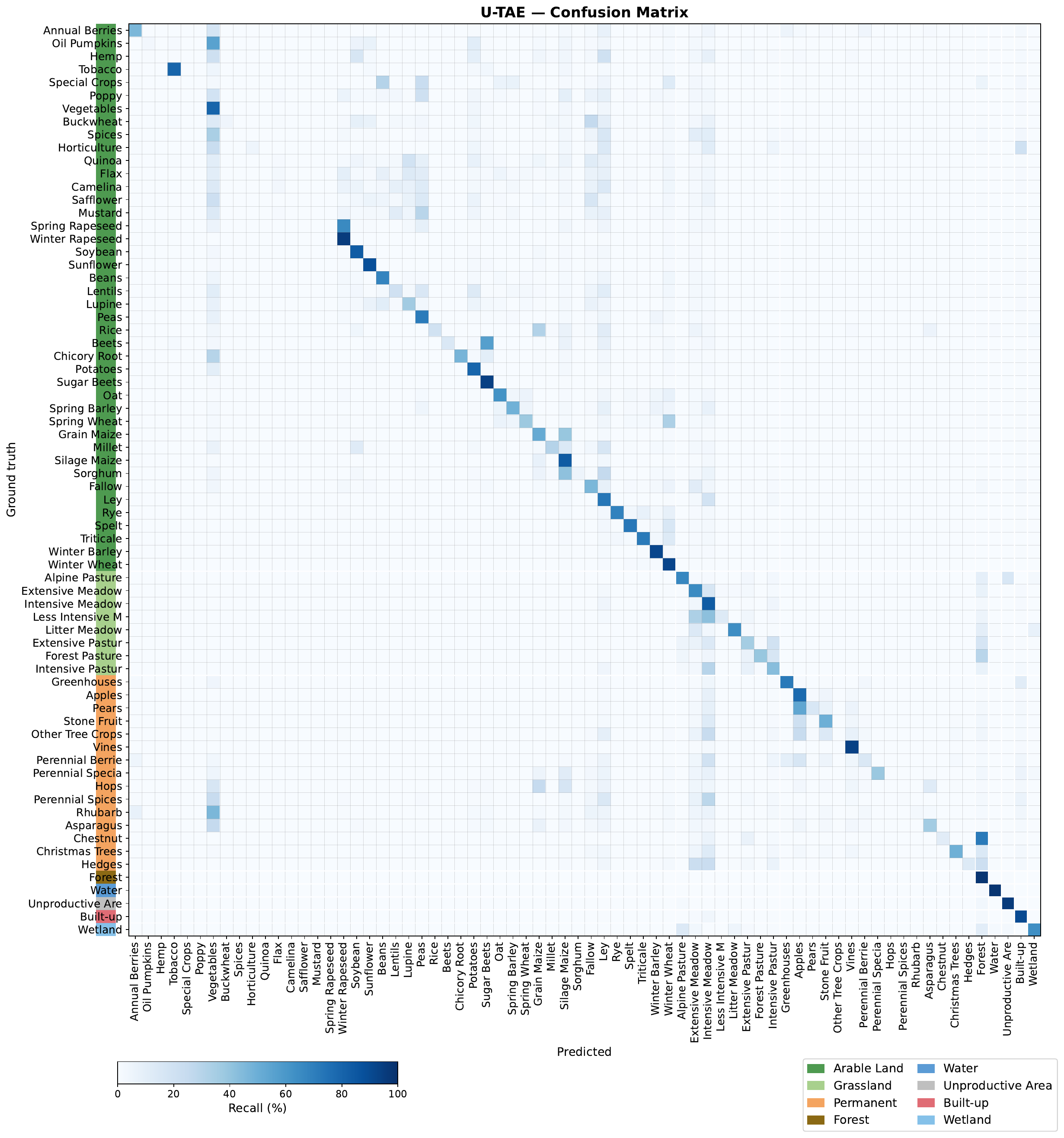}
  \caption{Row-normalised confusion matrix (recall, \%) for U-TAE, pooled across five LOYO splits. Classes are sorted by taxonomy group (colour bands on the left spine): Arable Land, Grassland, Permanent, Forest, Water, Unproductive Area, Built-up, Wetland. White lines mark group boundaries.}
  \label{supp:fig:confmat_utae}
\end{figure}

\begin{figure}[H]
  \centering
  \includegraphics[width=\linewidth]{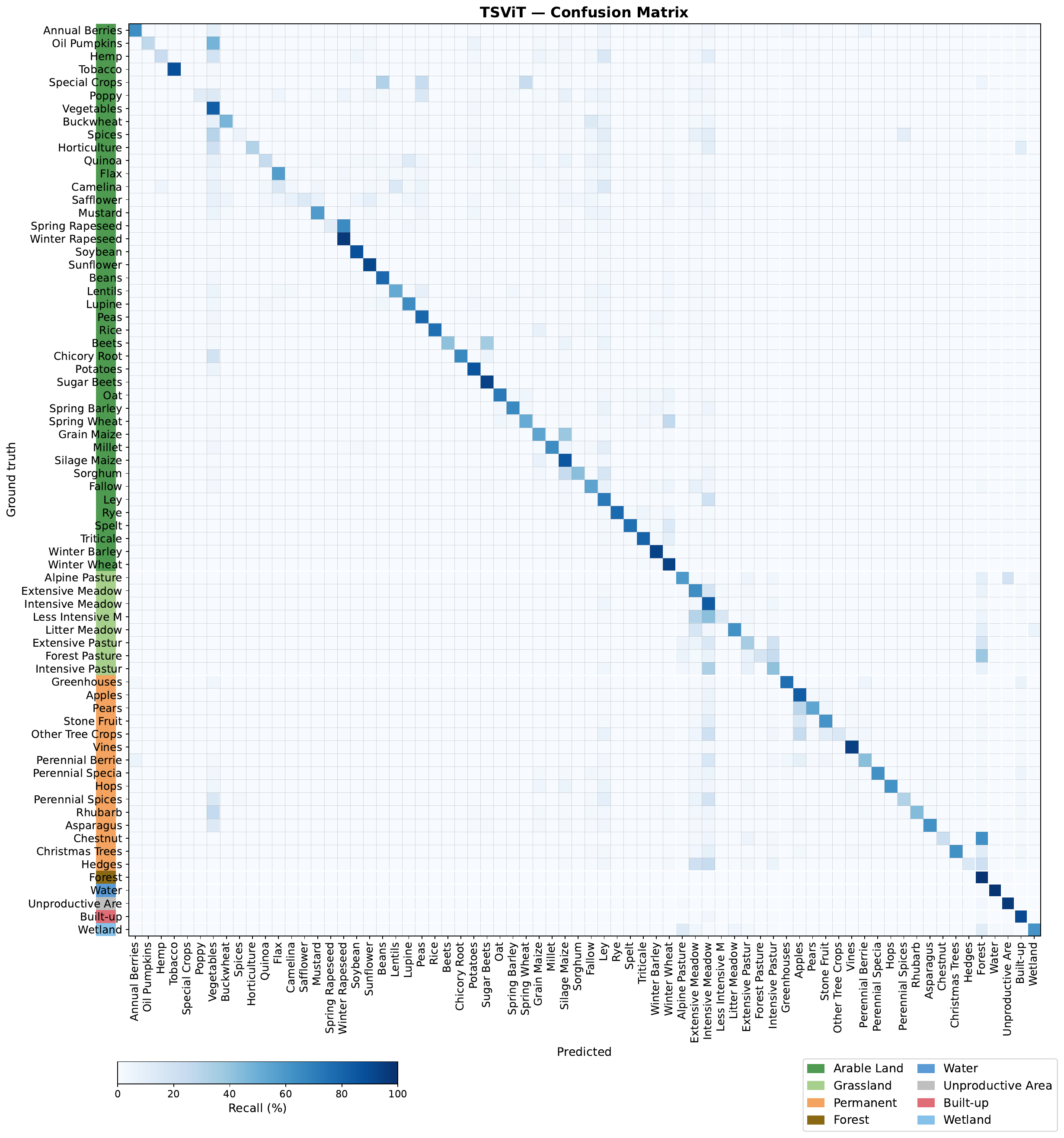}
  \caption{Row-normalised confusion matrix (recall, \%) for TSViT, pooled across five LOYO splits. Same class ordering and colour scheme as \cref{supp:fig:confmat_utae}.}
  \label{supp:fig:confmat_tsvit}
\end{figure}

\begin{figure}[H]
  \centering
  \includegraphics[width=\linewidth]{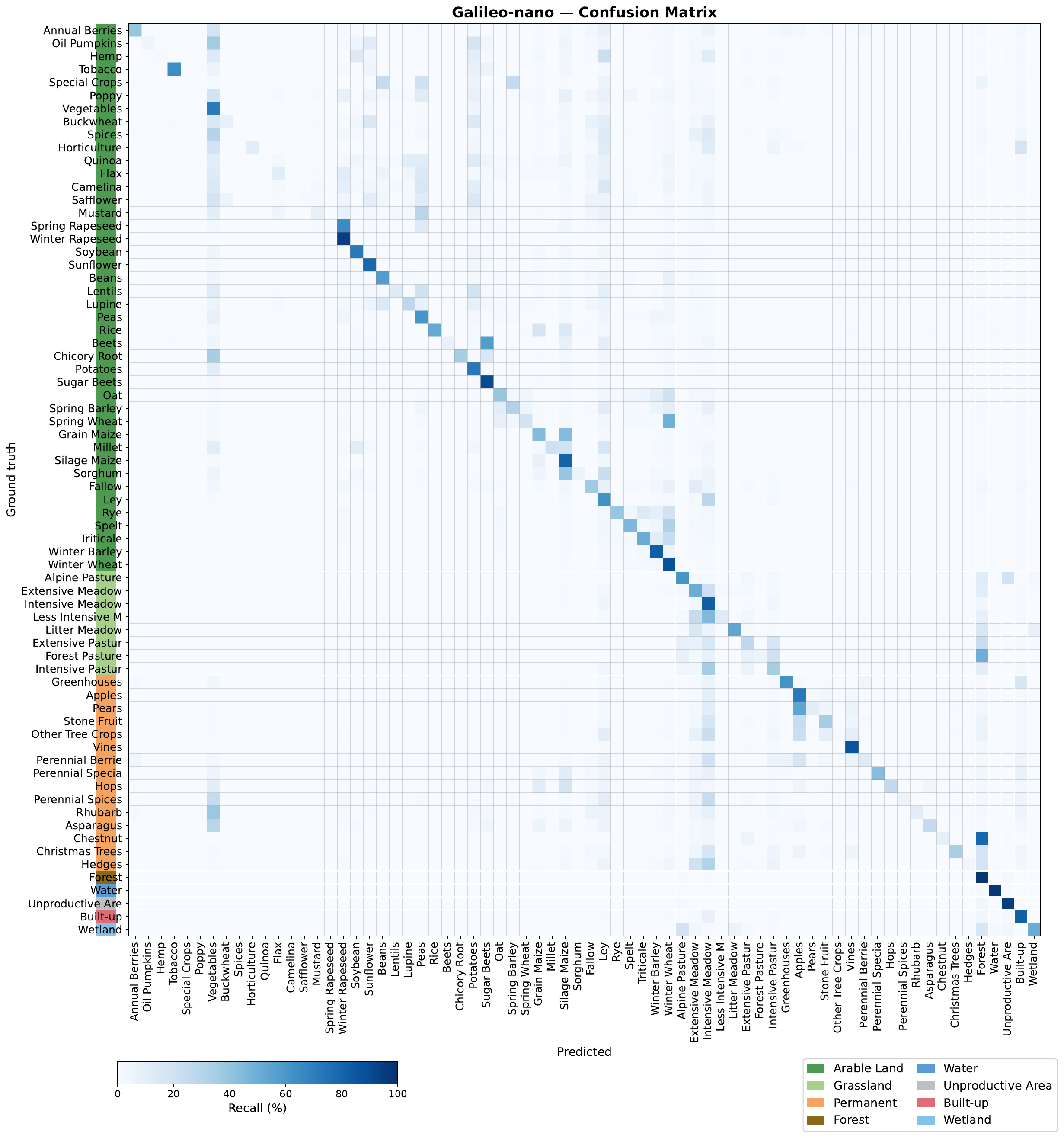}
  \caption{Row-normalised confusion matrix (recall, \%) for Galileo-nano, pooled across five LOYO splits. Same class ordering and colour scheme as \cref{supp:fig:confmat_utae}.}
  \label{supp:fig:confmat_galileo}
\end{figure}

\begin{figure}[H]
  \centering
  \includegraphics[width=\linewidth]{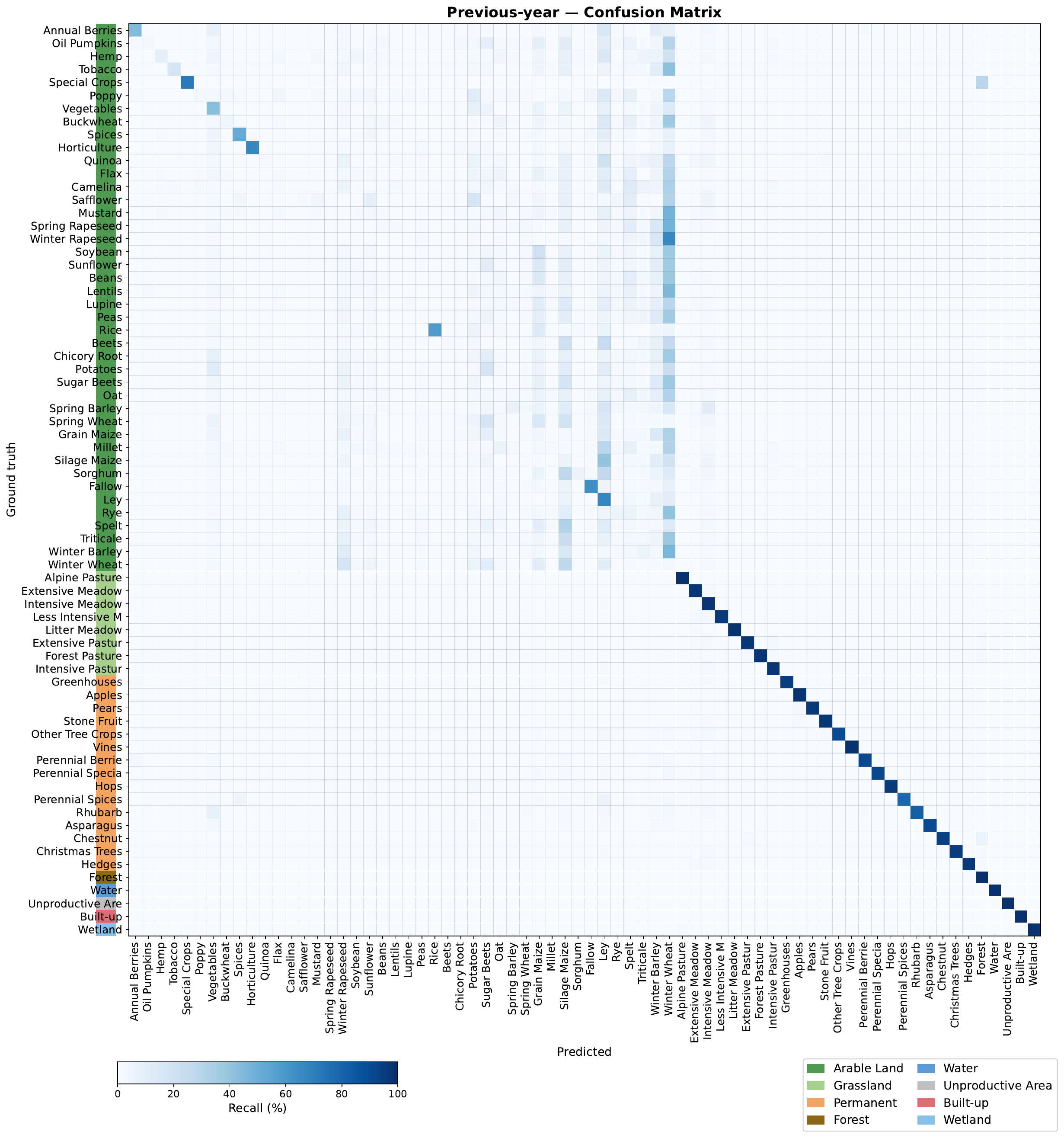}
  \caption{Row-normalised confusion matrix (recall, \%) for the previous-year naive temporal baseline, pooled across four LOYO splits (test years 2022--2025). Strong diagonal recall in the Grassland and Permanent blocks reflects year-to-year spatial stability; arable crop rows show near-zero recall due to crop rotation. Same class ordering and colour scheme as \cref{supp:fig:confmat_utae}.}
  \label{supp:fig:confmat_prev_year}
\end{figure}

\begin{figure}[H]
  \centering
  \includegraphics[width=\linewidth]{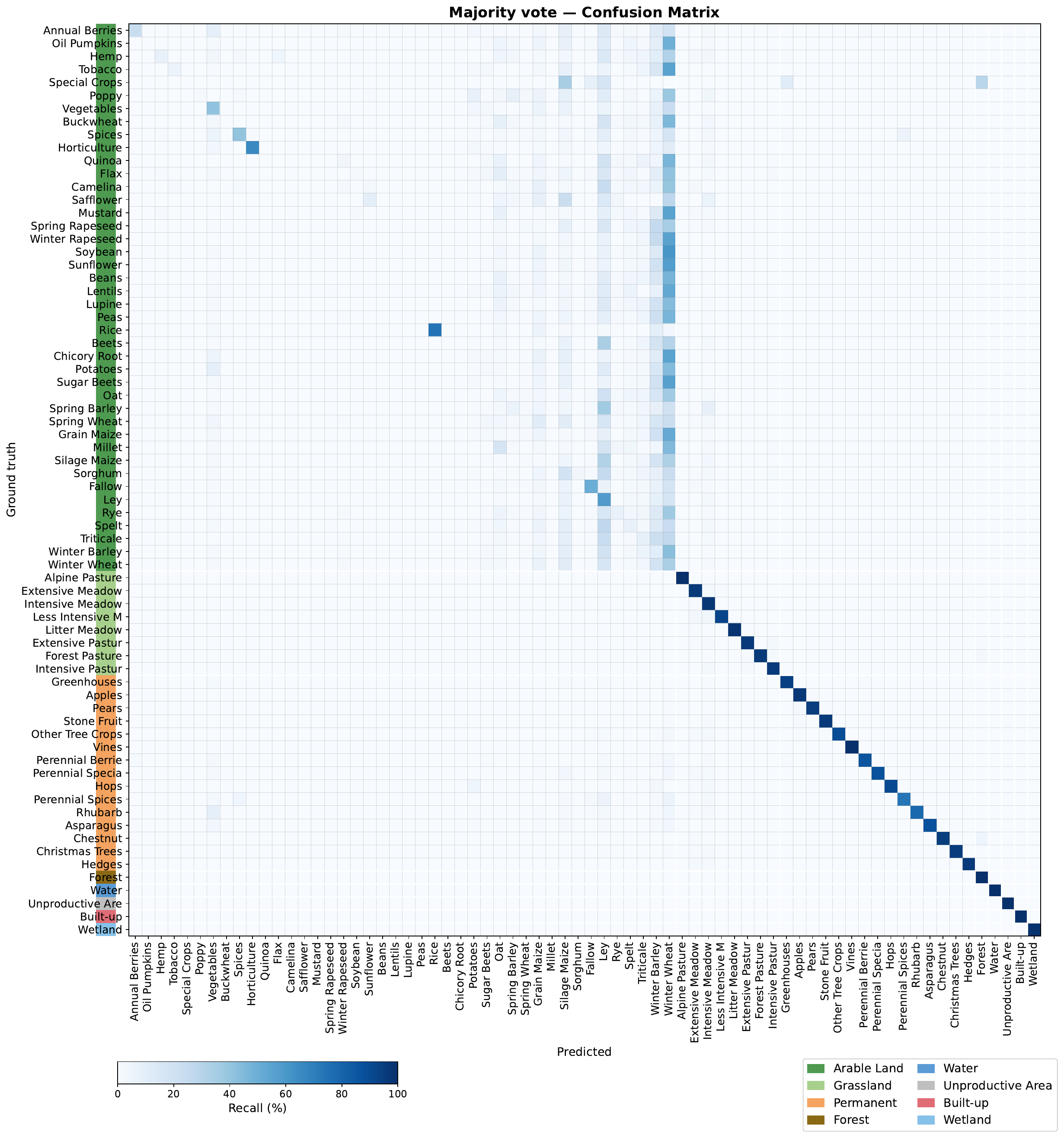}
  \caption{Row-normalised confusion matrix (recall, \%) for the majority-vote naive temporal baseline, pooled across four LOYO splits (test years 2022--2025). Strong diagonal recall in the Grassland and Permanent blocks reflects year-to-year spatial stability; arable crop rows show near-zero recall due to crop rotation. Same class ordering and colour scheme as \cref{supp:fig:confmat_utae}.}
  \label{supp:fig:confmat_majority}
\end{figure}

\newpage
\subsection*{Taxonomy granularity}

\begin{figure}[H]
  \centering
  \includegraphics[width=\linewidth]{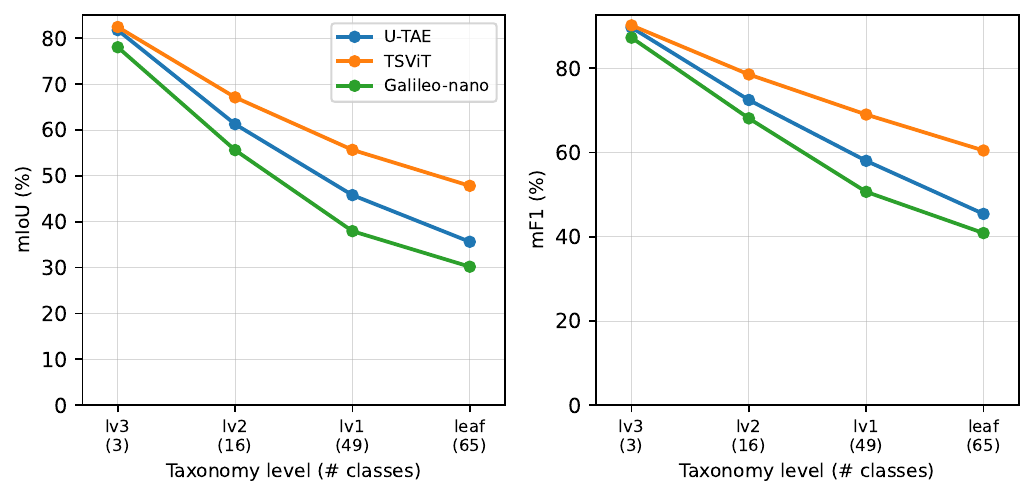}
  \caption{Crop mIoU (left) and mF1 (right) as a function of taxonomy granularity, from 3 coarse land-use categories (lv3) to the full \nrCropClasses-class operational leaf taxonomy. Only agricultural classes are included; land cover classes are evaluated separately in the main paper. At lv3, all models score within 5\,pp of each other in mIoU, but gaps widen sharply as agronomic specificity increases, showing that coarse evaluation hides meaningful architectural differences.}
  \label{supp:fig:taxonomy}
\end{figure}

\newpage
\subsection*{Long-tail difficulty}

\begin{figure}[H]
  \centering
  \includegraphics[width=\linewidth]{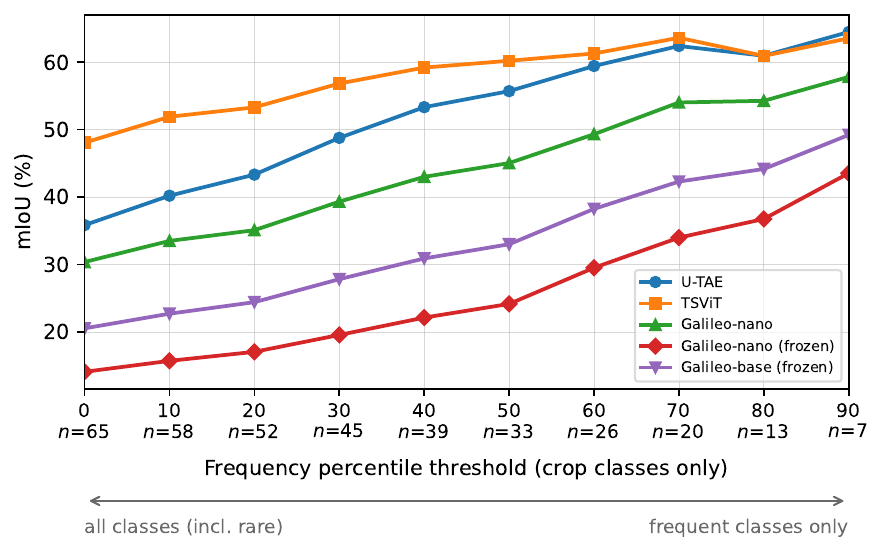}
  \caption{Crop mIoU of classes above a given test-frequency percentile threshold, averaged over all five splits. At threshold $T$, only the $(100-T)\%$ most frequent crop classes are included. The vertical spread between models widens towards the left (rare classes), quantifying long-tail difficulty.}
  \label{supp:fig:longtail}
\end{figure}


\newpage
\section{In-Season Usability}
\label{supp:inseason}

\subsection*{Evaluation protocol}
\label{supp:inseason_protocol}
All models are evaluated using the same weights and configurations as in the full-season benchmark; in-season evaluation only truncates the available time series after each monthly cutoff.
U-TAE zero-pads the time series to a fixed length and masks the padded positions in its temporal attention, so in-season evaluation simply applies the same mask to all future time steps beyond the monthly cutoff.
TSViT is evaluated with its variable-length variant, which natively accepts sequences of any length and does not require padding; truncation at a monthly cutoff is handled directly by reducing the sequence length.
Galileo-nano similarly supports variable-length inputs and is evaluated without padding.

\begin{figure}[H]
  \centering
  \includegraphics[width=.9\linewidth]{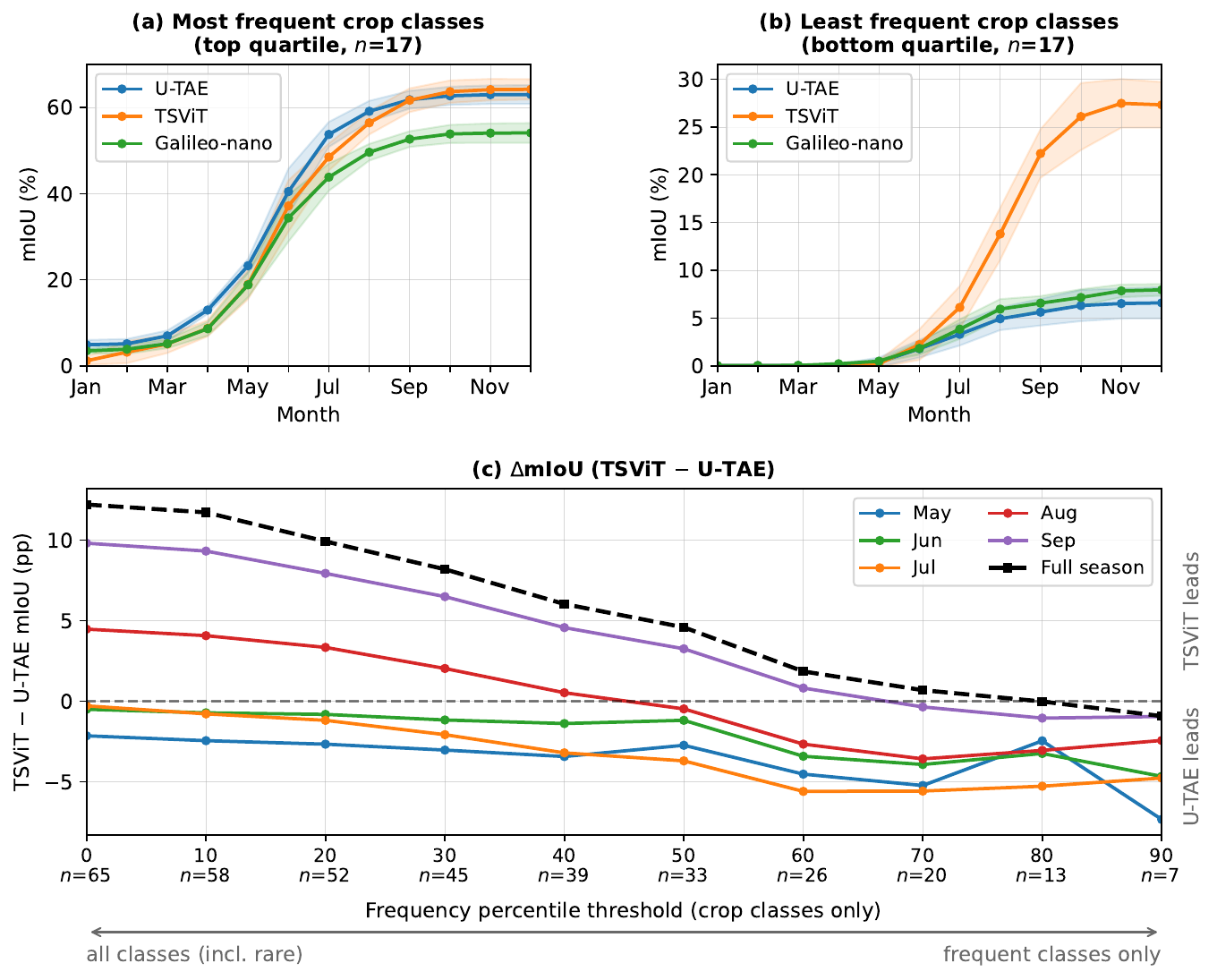}
  \caption{In-season mIoU stratified by class frequency.
    \textbf{(a)} Frequent crop classes (top quartile, $n=17$, ${\geq}8$M ground-truth pixels): U-TAE and TSViT track closely, with U-TAE retaining a small advantage until late season.
    \textbf{(b)} Rare crop classes (bottom quartile, $n=17$, ${\leq}147$K ground-truth pixels): TSViT gains a strong late-season advantage, reaching +20.7\,pp mIoU at the end of season. Galileo-nano and U-TAE follow near-identical trajectories throughout the season.
    Shaded bands show ${\pm}1$\,std across five LOYO splits.
    \textbf{(c)} $\Delta$mIoU (TSViT $-$ U-TAE) across frequency thresholds and months. At threshold $T$, only the $(100-T)\%$ most frequent classes are included ($n$ shown on x-axis). TSViT's advantage emerges first when rare classes are included and extends towards more common classes later in the season.}
  \label{supp:fig:inseason_classfreq}
\end{figure}


\newpage
\section{Efficiency and Scalability}
\label{supp:efficiency}

\subsection*{Low-resource evaluation}
\Cref{supp:tab:lowresource} extends the low-resource analysis from \cref{sec:exp:efficiency} with full per-split results.

\begin{table}[H]
\centering
\caption{Low-resource evaluation under the LOYO protocol: models trained on 10\% of available training data. Same metrics and protocol as \cref{tab:results}. Best value per split and metric is \textbf{bold}.}
\label{supp:tab:lowresource}
\setlength{\tabcolsep}{5pt}
\resizebox{\linewidth}{!}{
\begin{tabular}{llrrrrrr}
\toprule
\textbf{Test year} & \textbf{Model} & \textbf{OA (\%) $\uparrow$} & \textbf{GIoU (\%) $\uparrow$} & \textbf{mIoU (\%) $\uparrow$} & \textbf{mF1 (\%) $\uparrow$} & \textbf{ECE (\%) $\downarrow$} & \textbf{NLL\,(-) $\downarrow$} \\
\midrule
2021 & U-TAE (10\%) & 68.9 & 52.5 & 18.7 & 24.8 & 20.38 & 0.91 \\
 & TSViT (10\%) & \textbf{69.0} & \textbf{52.7} & \textbf{29.2} & \textbf{39.7} & 3.22 & \textbf{0.54} \\
 & Galileo-nano (10\%) & 65.6 & 48.8 & 17.1 & 24.3 & \textbf{0.94} & 0.61 \\
\midrule
2022 & U-TAE (10\%) & 70.7 & 54.7 & 18.0 & 23.5 & 19.42 & 0.83 \\
 & TSViT (10\%) & \textbf{72.1} & \textbf{56.3} & \textbf{33.3} & \textbf{44.5} & \textbf{0.30} & \textbf{0.48} \\
 & Galileo-nano (10\%) & 67.2 & 50.6 & 19.5 & 27.1 & 1.10 & 0.57 \\
\midrule
2023 & U-TAE (10\%) & 65.8 & 49.0 & 16.4 & 21.9 & 18.18 & 0.93 \\
 & TSViT (10\%) & \textbf{71.6} & \textbf{55.7} & \textbf{30.0} & \textbf{40.9} & 0.47 & \textbf{0.50} \\
 & Galileo-nano (10\%) & 64.3 & 47.4 & 17.4 & 24.6 & \textbf{0.17} & 0.61 \\
\midrule
2024 & U-TAE (10\%) & 64.9 & 48.0 & 16.0 & 21.5 & 18.32 & 0.94 \\
 & TSViT (10\%) & \textbf{68.4} & \textbf{52.0} & \textbf{27.9} & \textbf{38.6} & 1.27 & \textbf{0.56} \\
 & Galileo-nano (10\%) & 65.7 & 48.9 & 18.1 & 25.7 & \textbf{0.94} & 0.62 \\
\midrule
2025 & U-TAE (10\%) & 71.0 & 55.0 & 17.7 & 23.2 & 18.20 & 0.89 \\
 & TSViT (10\%) & \textbf{73.0} & \textbf{57.5} & \textbf{33.8} & \textbf{44.8} & 2.23 & \textbf{0.49} \\
 & Galileo-nano (10\%) & 68.6 & 52.2 & 20.0 & 27.7 & \textbf{0.38} & 0.57 \\
\midrule
\midrule
Mean $\pm$ Std & U-TAE (10\%) & 68.2 $\pm$ 2.8 & 51.9 $\pm$ 3.2 & 17.4 $\pm$ 1.1 & 23.0 $\pm$ 1.3 & 18.90 $\pm$ 0.98 & 0.90 $\pm$ 0.04 \\
 & TSViT (10\%) & \textbf{70.8} $\pm$ 2.0 & \textbf{54.8} $\pm$ 2.4 & \textbf{30.8} $\pm$ 2.6 & \textbf{41.7} $\pm$ 2.8 & 1.50 $\pm$ 1.23 & \textbf{0.51} $\pm$ 0.03 \\
 & Galileo-nano (10\%) & 66.3 $\pm$ 1.6 & 49.6 $\pm$ 1.8 & 18.4 $\pm$ 1.3 & 25.9 $\pm$ 1.5 & \textbf{0.70} $\pm$ 0.41 & 0.60 $\pm$ 0.02\\
\bottomrule
\end{tabular}
}
\end{table}

\subsection*{Model overview and computational cost}

\Cref{supp:tab:compute} reports model type, pretraining status, total and trainable parameter counts, and per-split training and inference times.

\begin{table}[H]
\centering
\caption{Model overview and computational cost per LOYO split. Training on 4 NVIDIA GH200 GPUs; inference on a single GH200.
  Inference time reflects a single pass over one complete year (${\sim}$26{,}000 tiles),
  equivalent to national-scale annual deployment.
  \emph{Type} describes the core architectural paradigm: convolutional encoder with temporal attention (Conv.+attn.), spatio-temporal transformer (Transformer), or Earth observation foundation model (FM).
  \emph{Pretrained} indicates large-scale Earth observation pretraining.
  \emph{Total} counts all model parameters; \emph{Trainable} counts parameters updated during fine-tuning (head only for frozen variants).}
\label{supp:tab:compute}
\setlength{\tabcolsep}{5pt}
\resizebox{\textwidth}{!}{%
\begin{tabular}{llcrrrrr}
\toprule
\textbf{Model} & \textbf{Type} & \textbf{Pretrained} & \textbf{Total (M)} & \textbf{Trainable (M)} & \textbf{Train (h) $\downarrow$} & \textbf{Inference (min) $\downarrow$} \\
\midrule
U-TAE                     & Conv.+attn.     & --- & 4.0 & 4.0 & 8.9 & 7.4 \\
TSViT                     & Transformer     & --- & 1.6 & 1.6 & 12.0 & 6.5 \\
Galileo-nano              & FM              & \checkmark & 1.9 & 1.9 & 36.2 & 15.9 \\
Galileo-nano (frozen)     & FM              & \checkmark & 1.9 & 0.9 & 10.7 & 9.7 \\
Galileo-base (frozen)     & FM              & \checkmark & 88.9 & 2.4 & 35.1 & 31.2\\
\bottomrule
\end{tabular}
}
\end{table}

\newpage
\section{Per-Class Results}
\label{supp:perclass}

\Cref{supp:tab:perclass} reports per-class IoU for all \nrCropClasses\ agricultural crop classes, averaged across all five LOYO splits.
Classes are grouped by taxonomy and sorted within each group by mean IoU across models (descending).
The best IoU per class is \textbf{bold}.
Landscape classes (forest, water, built-up, unproductive land, wetland) are excluded here and evaluated as a binary cropland mask in the main paper.

\begin{longtable}{lrrr}
\caption{Per-class IoU (\%) for all \nrCropClasses\ agricultural crop classes, averaged across five LOYO splits. Classes are grouped by taxonomy and sorted within each group by mean IoU (descending). Best IoU per row is \textbf{bold}.}
\label{supp:tab:perclass} \\
\toprule
\textbf{Class} & \textbf{U-TAE} & \textbf{TSViT} & \textbf{Galileo-nano} \\
 & IoU (\%) & IoU (\%) & IoU (\%) \\
\midrule
\endfirsthead
\multicolumn{4}{l}{\small\textit{(continued from previous page)}} \\
\toprule
\textbf{Class} & \textbf{U-TAE} & \textbf{TSViT} & \textbf{Galileo-nano} \\
 & IoU (\%) & IoU (\%) & IoU (\%) \\
\midrule
\endhead
\midrule
\multicolumn{4}{r}{\small\textit{(continued on next page)}} \\
\endfoot
\bottomrule
\endlastfoot
\textit{\textbf{Arable Land}} & \textbf{85.1} & 84.9 & 81.4 \\
\midrule
~~\textbf{Winter Cereals} & 89.1 & \textbf{89.7} & 84.1 \\
\midrule
~~~~~~Winter Wheat & 83.7 & \textbf{85.3} & 74.2 \\
~~~~~~Winter Barley & 82.3 & \textbf{84.4} & 67.3 \\
~~~~~~Spelt & 61.0 & \textbf{65.1} & 37.2 \\
~~~~~~Triticale & 56.9 & \textbf{62.4} & 35.3 \\
~~~~~~Rye & 55.4 & \textbf{67.6} & 28.6 \\
\addlinespace[3pt]
~~\textbf{Spring Cereals} & 52.3 & \textbf{59.4} & 30.0 \\
\midrule
~~~~~~Oat & 51.8 & \textbf{61.5} & 25.9 \\
~~~~~~Spring Barley & 34.1 & \textbf{48.3} & 21.6 \\
~~~~~~Spring Wheat & 29.8 & \textbf{40.9} & 13.2 \\
\addlinespace[3pt]
~~\textbf{Summer Cereals} & 82.3 & \textbf{83.5} & 77.7 \\
\midrule
~~~~~~Silage Maize & 67.0 & \textbf{68.6} & 62.2 \\
~~~~~~Grain Maize & 40.6 & \textbf{42.7} & 34.0 \\
~~~~~~Millet & 25.8 & \textbf{51.0} & 16.0 \\
~~~~~~Sorghum & 4.8 & \textbf{30.9} & 5.9 \\
\addlinespace[3pt]
~~\textbf{Oilseeds} & 86.3 & \textbf{88.7} & 78.8 \\
\midrule
~~~~\textit{Rapeseed} & 89.7 & \textbf{90.7} & 84.0 \\
~~~~~~~~Winter Rapeseed & 89.6 & \textbf{90.5} & 83.8 \\
~~~~~~~~Spring Rapeseed & 0.2 & \textbf{11.2} & 0.1 \\
\addlinespace[2pt]
~~~~~~Sunflower & 79.9 & \textbf{84.1} & 67.9 \\
~~~~~~Soybean & 70.7 & \textbf{78.8} & 55.6 \\
~~~~\textit{Minor Oilseeds} & 2.1 & \textbf{44.9} & 9.7 \\
~~~~~~~~Flax & 2.1 & \textbf{47.2} & 9.0 \\
~~~~~~~~Mustard & 0.0 & \textbf{43.7} & 7.3 \\
~~~~~~~~Safflower & 0.0 & \textbf{5.4} & 0.2 \\
~~~~~~~~Camelina & 0.0 & \textbf{1.2} & 0.0 \\
\addlinespace[3pt]
~~\textbf{Root Crops} & 81.2 & \textbf{84.8} & 73.6 \\
\midrule
~~~~~~Sugar Beets & 87.1 & \textbf{88.3} & 80.5 \\
~~~~~~Potatoes & 70.6 & \textbf{77.2} & 60.7 \\
~~~~~~Chicory Root & 39.9 & \textbf{52.0} & 27.4 \\
~~~~~~Beets & 12.8 & \textbf{26.1} & 6.2 \\
\addlinespace[3pt]
~~\textbf{Pulses} & 56.9 & \textbf{67.5} & 47.2 \\
\midrule
~~~~~~Peas & 58.0 & \textbf{67.4} & 45.9 \\
~~~~~~Beans & 56.3 & \textbf{67.9} & 45.9 \\
~~~~~~Lupine & 27.6 & \textbf{53.8} & 20.5 \\
~~~~~~Lentils & 15.4 & \textbf{39.0} & 10.3 \\
\addlinespace[3pt]
~~\textbf{Annual Horticulture} & 59.8 & \textbf{66.9} & 51.3 \\
\midrule
~~~~\textit{Vegetables} & 59.1 & \textbf{66.4} & 50.9 \\
~~~~~~~~Vegetables & 59.5 & \textbf{66.7} & 51.2 \\
~~~~~~~~Horticulture & 5.5 & \textbf{28.8} & 12.5 \\
~~~~~~~~Buckwheat & 3.7 & \textbf{35.2} & 6.5 \\
~~~~~~~~Quinoa & 0.0 & \textbf{12.3} & 0.0 \\
~~~~~~~~Spices & 0.0 & \textbf{4.5} & 0.0 \\
\addlinespace[2pt]
~~~~\textit{Special Crops} & 56.7 & \textbf{64.8} & 44.2 \\
~~~~~~~~Tobacco & 65.0 & \textbf{76.0} & 51.1 \\
~~~~~~~~Hemp & 0.0 & \textbf{17.7} & 0.7 \\
~~~~~~~~Poppy & 0.0 & \textbf{7.9} & 0.0 \\
~~~~~~~~Special Crops & \textbf{0.0} & 0.0 & 0.0 \\
\addlinespace[2pt]
~~~~~~Annual Berries & 38.1 & \textbf{49.9} & 32.1 \\
~~~~~~Oil Pumpkins & 2.5 & \textbf{23.8} & 3.9 \\
\addlinespace[3pt]
~~\textbf{Temporary Grassland} & \textbf{59.9} & 59.2 & 51.0 \\
\midrule
~~~~~~Ley & \textbf{60.2} & 59.1 & 51.2 \\
~~~~~~Fallow & 39.5 & \textbf{47.1} & 32.3 \\
\addlinespace[3pt]
~~\textbf{Rice} & 19.3 & \textbf{58.3} & 41.4 \\
\midrule
~~~~~~Rice & 19.3 & \textbf{58.3} & 41.4 \\
\addlinespace[6pt]
\textit{\textbf{Permanent}} & 68.7 & \textbf{71.2} & 63.5 \\
\midrule
~~\textbf{Perennial Horticulture} & 18.0 & \textbf{40.2} & 15.9 \\
\midrule
~~~~\textit{Perennial Special Crops} & 30.8 & \textbf{57.2} & 37.8 \\
~~~~~~~~Perennial Special Crops & 38.1 & \textbf{57.8} & 41.2 \\
~~~~~~~~Hops & 0.0 & \textbf{56.0} & 24.3 \\
\addlinespace[2pt]
~~~~\textit{Perennial Vegetables} & 24.5 & \textbf{48.5} & 19.1 \\
~~~~~~~~Asparagus & 28.2 & \textbf{50.3} & 20.6 \\
~~~~~~~~Rhubarb & 2.1 & \textbf{38.6} & 10.6 \\
\addlinespace[2pt]
~~~~~~Perennial Berries & 12.0 & \textbf{34.1} & 11.3 \\
~~~~~~Perennial Spices & 0.5 & \textbf{24.0} & 6.1 \\
\addlinespace[3pt]
~~\textbf{Orchard} & 84.7 & \textbf{87.0} & 79.5 \\
\midrule
~~~~~~Vines & 88.3 & \textbf{91.2} & 82.5 \\
~~~~\textit{Tree Crops} & 71.0 & \textbf{75.8} & 65.1 \\
~~~~~~~~Apples & 58.8 & \textbf{66.4} & 53.3 \\
~~~~~~~~Stone Fruit & 39.3 & \textbf{50.7} & 30.5 \\
~~~~~~~~Pears & 13.3 & \textbf{46.7} & 10.0 \\
~~~~~~~~Other Tree Crops & 0.1 & \textbf{14.5} & 1.2 \\
\addlinespace[3pt]
~~\textbf{Greenhouses} & 61.0 & \textbf{66.3} & 57.1 \\
\midrule
~~~~~~Greenhouses & 61.0 & \textbf{66.3} & 57.1 \\
\addlinespace[3pt]
~~\textbf{Woody Crops} & 17.6 & \textbf{20.7} & 8.0 \\
\midrule
~~~~~~Christmas Trees & 50.0 & \textbf{61.4} & 36.6 \\
~~~~~~Chestnut & 28.4 & \textbf{45.7} & 28.2 \\
~~~~~~Hedges & 13.3 & \textbf{14.7} & 4.0 \\
\addlinespace[6pt]
\textit{\textbf{Grassland}} & \textbf{91.5} & 91.3 & 89.3 \\
\midrule
~~\textbf{Meadow} & \textbf{76.4} & 75.7 & 72.0 \\
\midrule
~~~~~~Intensive Meadow & \textbf{68.4} & 67.4 & 64.3 \\
~~~~~~Litter Meadow & 63.1 & \textbf{64.1} & 56.0 \\
~~~~~~Extensive Meadow & 51.4 & \textbf{52.0} & 42.1 \\
~~~~~~Less Intensive Meadow & 11.6 & \textbf{13.0} & 10.9 \\
\addlinespace[3pt]
~~\textbf{Pasture} & \textbf{48.8} & 45.1 & 39.6 \\
\midrule
~~~~~~Intensive Pasture & \textbf{33.5} & 31.3 & 27.8 \\
~~~~~~Extensive Pasture & \textbf{29.8} & 28.2 & 24.2 \\
~~~~~~Forest Pasture & \textbf{41.5} & 23.9 & 11.7 \\
\addlinespace[3pt]
~~\textbf{Alpine Pasture} & \textbf{86.9} & 81.0 & 82.8 \\
\midrule
~~~~~~Alpine Pasture & \textbf{86.9} & 81.0 & 82.8 \\
\addlinespace[6pt]
\midrule
\midrule
\multicolumn{4}{l}{\textit{Mean IoU by taxonomy level}} \\
\midrule
\textbf{lv3 (3 groups)} & 81.8 & \textbf{82.5} & 78.1 \\
\textbf{lv2 (16 groups)} & 61.3 & \textbf{67.1} & 55.6 \\
\textbf{lv1 (49 groups)} & 45.8 & \textbf{55.7} & 38.0 \\
\textbf{leaf (65 classes)} & 35.8 & \textbf{48.1} & 30.4
\end{longtable}

\end{document}